\documentclass{article}

\usepackage[main,final]{neurips_2026}

\usepackage[utf8]{inputenc} %
\usepackage[T1]{fontenc}    %
\usepackage{hyperref}       %
\usepackage{amsmath}
\usepackage{cleveref}
\usepackage{url}            %
\usepackage{booktabs}       %
\usepackage{amsfonts}       %
\usepackage{nicefrac}       %
\usepackage{microtype}      %
\usepackage{xcolor}         %
\usepackage{graphicx}
\usepackage{enumitem}
\usepackage{wrapfig}
\usepackage{makecell}
\usepackage{bbm}
\usepackage{subcaption}
\usepackage{newtxtext}

\definecolor{meshbrown}{HTML}{B93030}
\definecolor{meshpink}{HTML}{BA03BB}
\definecolor{meshblue}{HTML}{467191}

\crefname{figure}{Fig.}{Figs.}
\Crefname{figure}{Fig.}{Figs.}
\crefname{section}{Sec.}{Secs.}
\Crefname{section}{Sec.}{Secs.}

\usepackage[table]{xcolor}
\usepackage{booktabs}
\usepackage{multirow}

\definecolor{sectionblue}{RGB}{220,239,246}
\definecolor{methodpink}{RGB}{242,231,239}
\definecolor{refgray}{gray}{0.48}
\newcommand{\gray}[1]{\textcolor{refgray}{#1}}
\newcommand{\sectionrow}[1]{%
  \multicolumn{13}{@{}l@{}}{%
    \cellcolor{sectionblue}%
    \makebox[\dimexpr\linewidth\relax][l]{\textit{#1}}%
  }\\
}

\newcommand{\change}[1]{{\color{red}{#1}}}
\renewcommand{\change}[1]{{#1}}

\newcommand{\methodName}{\textsc{PhyMotion}}

\title{\textsc{\methodName}: Structured 3D Motion Reward for Physics-Grounded Human Video Generation}

\author{
\textbf{Yidong Huang$^{1}$\thanks{Equal contribution.} \quad
Zun Wang$^{1}$\footnotemark[1] \quad
Han Lin$^{1}$ \quad
Dong-Ki Kim$^{2}$ \quad
Shayegan Omidshafiei$^{2}$} \\
\textbf{Jaehong Yoon$^{3}$ \quad
Jaemin Cho$^{4,5}$ \quad
Yue Zhang$^{1}$ \quad
Mohit Bansal$^{1}$ \vspace{8pt} }\\
$^{1}$UNC Chapel Hill \quad
$^{2}$FieldAI \quad
$^{3}$NTU Singapore \quad 
$^{4}$AI2 \quad $^{5}$Johns Hopkins University
\vspace{8pt} \\
{\tt \href{https://phy-motion.github.io/}{\textbf{https://phy-motion.github.io/}}}
}

\begin{document}

\maketitle

\begin{abstract}
Generating realistic human motion is a central yet unsolved challenge in video generation. 
While reinforcement learning (RL)-based post-training has driven recent gains in general video quality, extending it to human motion remains bottlenecked by a reward signal that cannot reliably score motion realism.
Existing video rewards primarily rely on 2D perceptual signals, without explicitly modeling the 3D body state, contact, and dynamics underlying articulated human motion, and often assign high scores to videos with floating bodies or physically implausible movements.
To address this, we propose \methodName{}, a structured, fine-grained motion reward that grounds recovered 3D human trajectories in a physics simulator and evaluates motion quality along multiple dimensions of physical feasibility. Concretely, we recover SMPL body meshes from generated videos, retarget them onto a humanoid in the MuJoCo physics simulator, and evaluate the resulting motion along three axes: kinematic plausibility, contact and balance consistency, and dynamic feasibility. Each component provides a continuous and interpretable signal tied to a specific aspect of motion quality, allowing the reward to capture which aspects of motion are physically correct or violated.
Experiments show that \methodName{} achieves stronger correlation with human judgments than existing reward formulations. These gains carry over to RL-based post-training, where optimizing \methodName{} leads to larger and more consistent improvements than optimizing existing rewards, improving motion realism across both autoregressive and bidirectional video generators under both automatic metrics and blind human evaluation (+68 Elo gain). Ablations show that the three axes provide complementary supervision signals, while the reward preserves overall video generation quality with only modest training overhead.

\end{abstract}
\section{Introduction}
\label{sec:introduction}

Recent video generation models~\citep{ho2022videodiffusion,openai2024sora,kong2024hunyuanvideo,wan2025wan} produce increasingly photorealistic outputs, yet generating realistic human motion remains one of the most significant open challenges. 
While scene textures, lighting, and camera motion have reached impressive fidelity~\citep{zheng2025vbench2,liu2023evalcrafter,gao2025seedance}, modeling human motion remains fundamentally challenging due to the tight coupling between body articulation and physical constraints.
As a result, generated videos frequently exhibit artifacts like floating feet, limb penetrations, anatomical distortions, and movements that defy basic laws of physics~\citep{zheng2025vbench2,yang2025echomotion}. As human-centric content constitutes one of the most important application domains, from entertainment and education to virtual communication~\citep{zhu2023human,sui2026survey}, improving human motion quality is a pressing goal.

 \begin{figure}
    \centering
\includegraphics[width=0.98\linewidth]{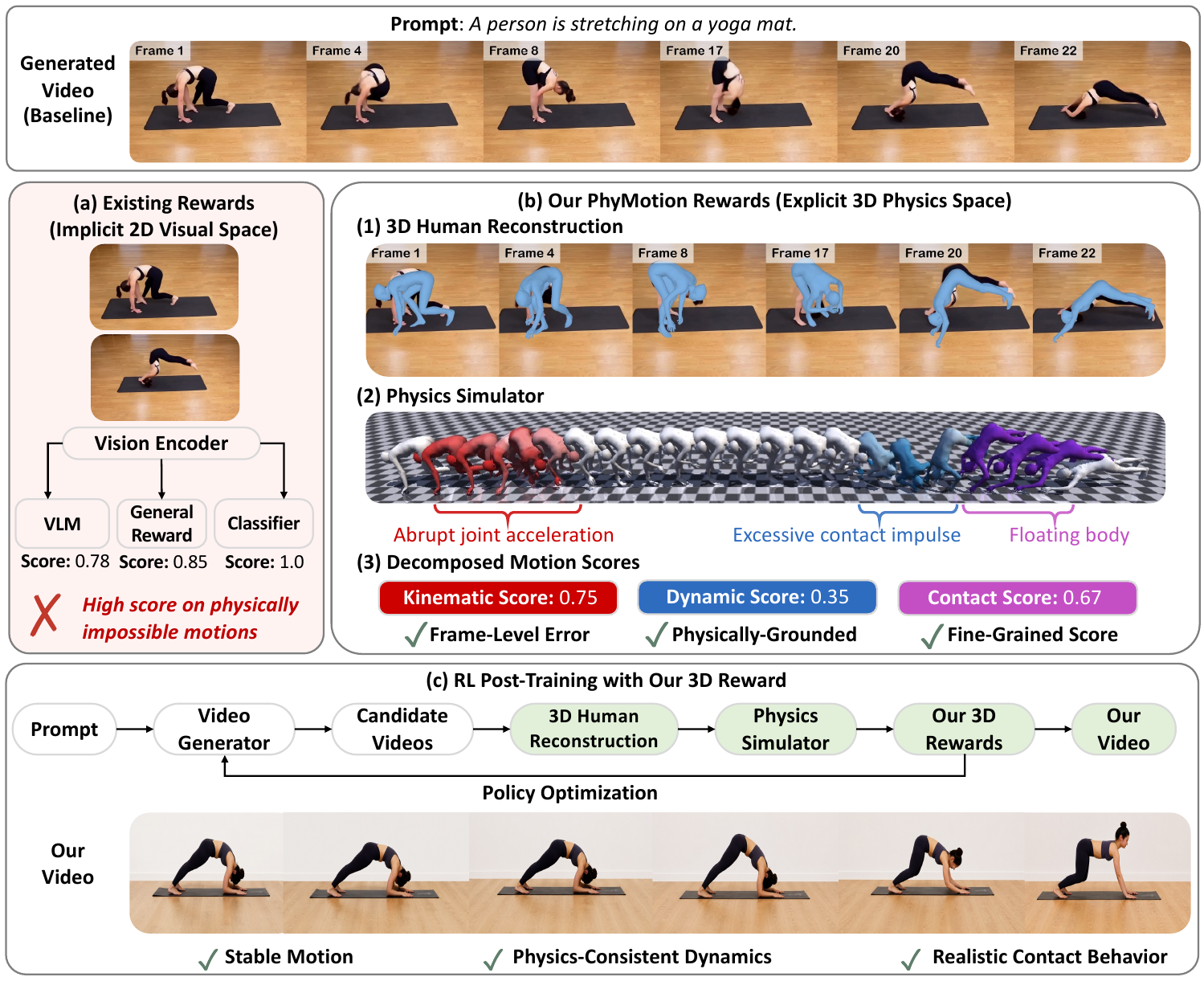}
    \caption{\textbf{Overview of \methodName.}  
    Existing metrics operate in 2D pixel space (VLMs~\citep{liu2025videoalign}, frame-level classifiers~\citep{zheng2025vbench2}), producing misleading scores that are structurally blind to articulated motion.  We recover SMPL meshes from generated videos and place them in a physics simulator, yielding decomposed, physically-grounded motion scores. The structured reward drives RL-based post-training (bottom), consistently improving human motion quality.
    }
    \label{fig:teaser}
    \vspace{-3mm}
\end{figure}

A promising direction is reinforcement learning (RL)-based post-training, which has recently demonstrated clear gains on general visual quality and text alignment~\citep{black2024ddpo,xue2025dancegrpo,zhang2026astrolabe}. 
The success of RL, however, hinges on the reward signal, and for human motion, existing rewards remain inadequate.
Specifically, they broadly fall into three categories, each with a distinct failure mode:  
\emph{(i) frame-level 2D classifiers}~\citep{huang2024vbench,zheng2025vbench2,motamed2026generative} detect localized artifacts but miss trajectory-level failures;
\emph{(ii) vision-language evaluators}~\citep{meng2024phygenbench,chow2025physbench,hong2025motionbench} provide coarse semantic judgments unsuitable as dense RL rewards; and
\emph{(iii) general preference rewards}~\citep{bansal2024videophy,liu2025videoalign,he2024videoscore,ma2025hpsv3} can hallucinate quality from appearance or prompt alignment rather than motion plausibility.
Despite their differences, these rewards primarily rely on pixel-space perceptual representations and learned visual features. While effective for appearance quality and semantic alignment, we empirically observe that they still struggle to reliably capture key aspects of articulated human motion and physical feasibility, such as kinematic consistency, anatomical constraints, and contact dynamics (see \Cref{sec:failure}). Consequently, they can assign high scores to videos with clear physical violations (\cref{fig:teaser}(a) and \cref{fig:metric_examples}).

To address this limitation, we propose \methodName, a structured, fine-grained motion reward for human video generation that grounds 3D human trajectories in a physics simulator to evaluate motion quality across multiple dimensions of physical feasibility (\cref{fig:teaser}(b)).
We lift each generated video into a structured 3D representation by recovering an SMPL~\citep{loper2015smpl} body mesh and grounding it in a physically consistent human model in the MuJoCo simulator~\citep{todorov2012mujoco}. This enables access to structured physical signals (e.g., contact consistency, joint-level kinematics, and motion-driving torques) that are not explicitly captured in pixel-based representations, allowing evaluation based on whether the motion satisfies physical constraints, which directly underlie many visually implausible artifacts.
From these observables, we derive three structured feasibility scores: \textbf{kinematic} (joint consistency), \textbf{contact/balance} (interaction correctness), and \textbf{dynamic} (force and motion consistency). Each score targets a distinct failure mode identified in \cref{sec:failure}. 
To test whether our physics-grounded scores better capture perceived motion quality, we conduct a pairwise human study in which raters compare pairs of generated videos according to motion realism, and measure how well each metric predicts these preferences. We compare \methodName{} against prior 2D perceptual metrics and learned video reward models. We further use the same scores as a dense and interpretable reward for RL-based post-training (\cref{fig:teaser}(c)), examining whether optimizing physical feasibility leads to improved human motion generation. Because the reward decomposes into physically meaningful components, it provides transparency into which aspects of motion are optimized, enabling fine-grained diagnosis of motion failures and revealing improvements across different physical dimensions.

Experiments demonstrate that \methodName{} serves as both a reliable motion evaluator and an effective reward for RL-based video post-training.
As an evaluator, \methodName{} achieves $80\%$ average pairwise agreement with human judgments and the highest aggregate Spearman correlation ($\rho{=}0.376$) on $1{,}200$ video pairs, across multiple aspects of motion quality including body structure, balance, and motion naturalness.
This substantially outperforms existing perceptual, preference-based, and physics-aware reward baselines, which typically achieve $50$--$66\%$ agreement and only weak Spearman correlations ($\rho=0$--$0.25$).
As a reward, \methodName{} further enables effective RL-based post-training across both autoregressive~\citep{zhu2026causal} and bidirectional~\citep{zhang2025fast} video generators.
Additionally, our reward also improves scores on external evaluators, including VBench metrics~\citep{huang2024vbench}, VideoAlign~\citep{liu2025videoalign}, and VideoPhy-PC~\citep{bansal2024videophy}, by an average of $7.1\%$, while achieving consistent gains across all three physical feasibility dimensions.
Human preference eval further confirms these improvements: our post-trained model achieves the highest Elo scores on body structure, balance, motion naturalness, and overall preference, outperforming all baselines, including the larger Wan2.2 14B model.
Meanwhile, our ablation studies show that each reward component improves its corresponding physical dimension, while the combined reward achieves the best overall trade-off across all dimensions. Finally, we further show that training with \methodName{} preserves general video generation quality on VBench and VBench-2.0~\citep{huang2024vbench,zheng2025vbench2}, while introducing only modest overhead through pipelined reward computation.

\section{Related Work}
\label{sec:related_work}

\noindent\textbf{RL-based post-training and video reward models.}
Reinforcement learning has become a central paradigm for post-training generative models~\citep{ouyang2022training,christiano2017rlhf,shao2024deepseekmath,guo2025deepseek}. For diffusion, DDPO~\citep{black2024ddpo}, DPOK~\citep{fan2024dpok}, and Diffusion-DPO~\citep{wallace2024diffusion} optimize policy gradients or preference objectives on reverse sampling, while DanceGRPO~\citep{xue2025dancegrpo}, Flow-GRPO~\citep{liu2025flowgrpo}, and DiffusionNFT~\citep{zheng2025diffusionnft} extend on-policy RL to video generators; recent work further scales RL to distilled autoregressive video models~\citep{zhang2026astrolabe,wang2026worldcompass,lu2025rewardforcing,he2025gardo}. The reward itself is typically a learned scalar over general human preferences, e.g., HPSv3~\citep{ma2025hpsv3}, VideoReward~\citep{liu2025videoalign}, VisionReward~\citep{xu2024visionreward}, and VideoScore~\citep{he2024videoscore}, or an aggregate of VBench-derived feature-model scores~\citep{huang2024vbench,huang2024vbench++}.
Such rewards are structurally agnostic to articulated human motion and provide no dimension-specific feedback.
In contrast, we derive the reward from a structured 3D body model, enabling the RL optimization signal to incorporate explicit physical priors on human kinematics, balance, and dynamics.

\noindent\textbf{Human motion in video generation and evaluation.}
A growing line of work improves human motion by directly modifying the generator
or incorporating motion priors. Pose-conditioned methods~\citep{wang2024disco,ma2024follow,xu2024magicanimate,hu2024animate,zhou2024realisdance,zhu2024champ,shao2024human4dit,wang2025epic,wang2026anchorweave}
condition on external 2D poses or 3D SMPL priors, while VideoJAM~\citep{chefer2025videojam}
and EchoMotion~\citep{yang2025echomotion} co-predict motion representations with video.
However, these methods typically require architectural changes, external pose
conditioning, or additional motion data supervision.
On the evaluation side, beyond distribution-level statistics
~\citep{salimans2016inceptionscore,heusel2017fid,unterthiner2019fvd}, recent
benchmarks~\citep{huang2024vbench,huang2024vbench++,zheng2025vbench2,liu2023evalcrafter,liu2023fetv,ling2025vmbench}
decompose evaluation into interpretable dimensions, but still rely largely on
perceptual video cues, providing only indirect evidence of articulated motion
quality. VLM-based evaluators
~\citep{bansal2024videophy,huang2025planning,meng2024phygenbench,wang2024storyeval,sun2024t2vcompbench,hong2025motionbench}
probe physical commonsense and motion understanding, but remain limited for
fine-grained human articulation. Related 3D motion evaluators such as
MBench~\citep{lin2025vimogen} and MotionCritic~\citep{wang2024motioncritic}
assess motion quality directly, but are not designed as rewards for synthetic videos.
Our work connects generation and evaluation through a structured 3D motion reward for human video generation. Without modifying the generator architecture, we use PhyMotion for RL post-training, providing continuous, physics-grounded supervision that targets distinct motion failure modes.

\section{Physics-Grounded 3D Motion Evaluation}

\begin{figure}[t]
    \centering
    \includegraphics[width=\linewidth]{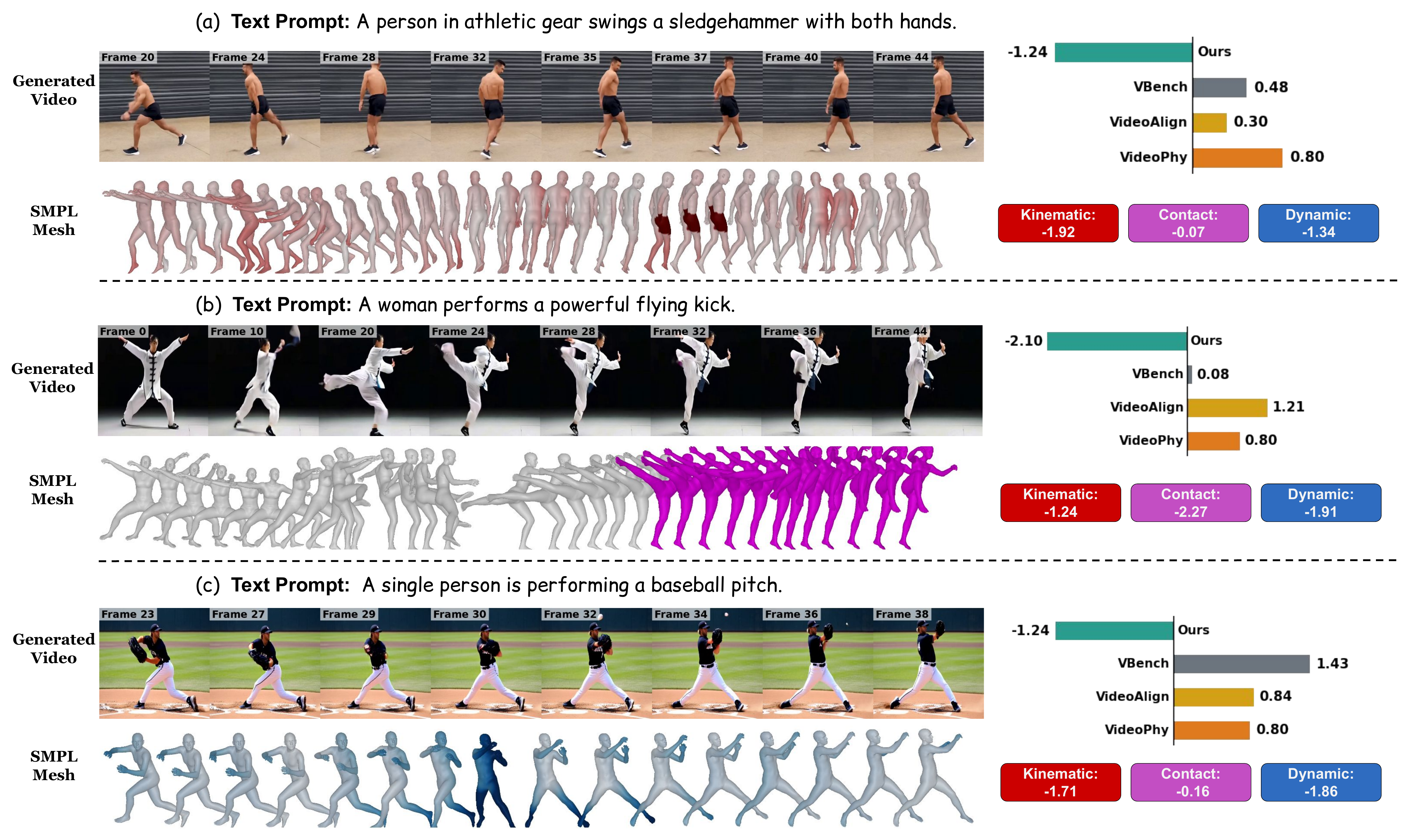}
\caption{
\textbf{\methodName{} identifies distinct physical failure modes overlooked by existing 2D metrics.}
Each example shows a generated video and its recovered SMPL trajectory. Colored SMPL
frames highlight the dominant motion errors: kinematic (articulated-body inconsistency), contact
(unstable body--environment interaction), and dynamic (physically infeasible motion).
All reported metric values are quartile-normalized z-scores.}
    \label{fig:metric_examples}
\end{figure}
\label{sec:evaluation}

We present our evaluation in two parts: we first characterize the failure modes of human motion that are fundamentally unidentifiable from pixel observations (\Cref{sec:failure}). We then introduce a physics-grounded evaluation protocol that recovers 3D motion along three feasibility axes (\Cref{sec:Physics-Grounded Evaluation}).

\subsection{The Failure Modes of 2D Motion Evaluation}
\label{sec:failure}
\Cref{fig:metric_examples} illustrates a key limitation of existing 2D motion
evaluators, such as VBench~\citep{huang2024vbench,zheng2025vbench2}, VideoAlign~\citep{liu2025videoalign}, and VideoPhy~\citep{bansal2024videophy}. 
Although these metrics can
assign high scores based on perceptual quality, text alignment, or visual
commonsense, the generated videos may still contain physically implausible human
motion.
These failures require reasoning about human bodies' physical quantities beyond RGB appearance:
whether the 3D body configuration is anatomically valid over time, whether the
body is properly supported by ground contacts, and whether the motion can be
produced by plausible forces and torques. 
Therefore, we identify three classes of motion failures that existing 2D evaluators struggle to reliably capture.

\noindent\textbf{Kinematic inconsistency over time.}
Frame-level visual metrics may judge individual poses as plausible, but fail to detect whether the recovered body configuration is anatomically valid in 3D.
Such metrics can
miss abnormal joint configurations, unrealistic joint velocities, and
self-penetrations that become clear only after reconstructing the articulated body.
As shown in \Cref{fig:metric_examples}(a), although the frames appear to show
a swing, the reconstructed 3D body reveals that the hand
penetrates the hip, a self-penetration error that existing 2D rewards fail to detect (e.g., VideoPhy assigns a high normalized z-score of $0.8$).

\noindent\textbf{Physically inconsistent contact.} 
2D evaluators often score a video high even when the body is floating, sliding,
penetrating the ground, losing balance, or changing contact state inconsistently
over time. As shown in \Cref{fig:metric_examples}(b), the RGB frames appear to show a
powerful flying kick, but the recovered 3D trajectory reveals an implausible
support pattern: the body becomes airborne toward the end of the video. Such contact and
balance failures are difficult for 2D rewards to detect (e.g. VideoAlign assigns a high normalized z-score of $1.21$).

\noindent\textbf{Dynamically infeasible motion.}
A video can appear smooth and semantically correct while still requiring
physically implausible forces to execute. As shown in
\Cref{fig:metric_examples}(c), the frames depict a reasonable baseball pitch,
but the recovered 3D motion indicates that the body would require excessive
joint torques to reproduce the observed pose transition. Such dynamic failures
are difficult for existing 2D metrics to detect because they do not test whether
the motion can be produced by a physically valid human body (e.g. VBench assigns a high normalized z-score of $1.43$).

\subsection{\methodName{}: Physics-Grounded 3D Rewards}
\label{sec:Physics-Grounded Evaluation}
To address the failure modes in \Cref{sec:failure}, we convert each generated
video into a physically interpretable 3D motion representation. Given a video
$\{I_t\}_{t=1}^{T}$ at frame rate $f$, we recover an SMPL-X~\citep{pavlakos2019smplx}
trajectory with GVHMR~\citep{shen2024gvhmr}. We denote the recovered pose by
$\mathbf{q}_t$, the 3D body joints by
$\mathbf{X}_t=\{\mathbf{x}_{t,j}\}_{j=1}^{J}$, and the joint angular velocity by
$\mathbf{v}_{t,j}=f(\mathbf{q}_{t+1,j}-\mathbf{q}_{t,j})$. We retarget the
motion to a MuJoCo~\citep{todorov2012mujoco} human model with explicit mass,
inertia, joint limits, and contact geometry, and run inverse
dynamics~\citep{winter2009biomechanics} to estimate joint torques
$\boldsymbol{\tau}_t$ and ground reaction forces $\mathbf{F}^{\mathrm{GRF}}_t$.
We then score each video along three complementary axes. This conversion makes the failure modes in \Cref{sec:failure} directly
measurable. As shown in \Cref{fig:metric_examples}, the recovered 3D trajectory lets us identify whether a
video fails due to kinematic inconsistency, implausible contact and balance, or
dynamically infeasible motion.

\noindent\textbf{Kinematic feasibility.}
Kinematic feasibility measures whether the recovered body motion is smooth and
anatomically valid. We combine three normalized violations: angular-velocity
violation $v_{\mathrm{vel}}$, computed by thresholding
$\|\mathbf{v}_{t,j}\|_2$ against a clean-motion tolerance; self-penetration
violation $v_{\mathrm{spen}}$, computed as the fraction of frames with
intersecting non-adjacent mesh triangles; and joint-limit violation
$v_{\mathrm{lim}}$, computed as the fraction of joints whose angles fall outside
the valid MuJoCo range. The final score is
$\mathcal{F}_{\mathrm{kin}}=1-\frac{1}{3}(v_{\mathrm{vel}}+v_{\mathrm{spen}}+v_{\mathrm{lim}})$,
where higher values indicate smoother motion with fewer anatomical violations.

\noindent{\textbf{Contact feasibility.}
Contact feasibility measures whether the body interacts with the ground
plausibly. We infer binary foot--ground contacts $c_{t,k}\in\{0,1\}$ for each
foot $k\in\{L,R\}$ from foot height and velocity, where $L$ and $R$ denote the
left and right foot. We compute four normalized violations: foot sliding
$v_{\mathrm{slip}}$,
which measures displacement of a contacted foot; ground penetration
$v_{\mathrm{gpen}}$, which measures how far the foot moves below the floor;
foot floating $v_{\mathrm{float}}$, which flags frames where neither foot is in
contact while the body does not follow a plausible ballistic trajectory; and
balance violation $v_{\mathrm{bal}}$ is the fraction of frames where the projected
center of mass falls outside the support polygon of contacting feet. The score
is
$\mathcal{F}_{\mathrm{con}}=1-\frac{1}{4}(v_{\mathrm{slip}}+v_{\mathrm{gpen}}+v_{\mathrm{float}}+v_{\mathrm{bal}})$.

\noindent{\textbf{Dynamic feasibility.}
Dynamic feasibility measures whether the recovered motion can be replayed by a
physically plausible human body. Using inverse dynamics in MuJoCo, we estimate
the forces required to reproduce the trajectory and compute three scores:
$s_{\boldsymbol{\tau}}$ penalizes unrealistically large joint torques,
$s_{\mathrm{GRF}}$ penalizes excessive ground contact forces whose magnitude
exceeds a maximum plausible threshold $F_{\max}$, and $s_{\mathrm{met}}$ penalizes motions with unusually high joint effort, measured
by the torque--velocity work proxy
$C_{\mathrm{met}}=\sum_{t,j}|\tau_{t,j}\mathbf{v}_{t,j}|$. The final score is
$\mathcal{F}_{\mathrm{dyn}}=\frac{1}{3}(s_{\boldsymbol{\tau}}+s_{\mathrm{GRF}}+s_{\mathrm{met}})$, with higher values indicating more physically realizable motion.

Together, these axes cover articulation quality, environment interaction, and
physical realizability, producing continuous and interpretable metrics for both
evaluation and reward-based post-training. We provide the detailed reward
definitions and implementation specifications in Appendix~\ref{app:metric_details}.

\section{RL Post-Training with \methodName}
\label{sec:rl}

The evaluation metrics in \Cref{sec:Physics-Grounded Evaluation} provide decomposed
signals for different human motion failure modes. To use them as an optimization
signal, we aggregate the three \methodName{}  scores into a single
motion reward:
\begin{equation}
\label{eq:reward-agg}
R_{\mathrm{motion}}(v)
=
\frac{1}{3}
\left(
\mathcal{F}_{\mathrm{kin}}(v)
+
\mathcal{F}_{\mathrm{con}}(v)
+
\mathcal{F}_{\mathrm{dyn}}(v)
\right).
\end{equation}
This reward is then used as the optimization target in the policy-learning objective below.

\noindent{\textbf{Problem formulation.}}
We frame the video generation model as a policy $\pi_\theta$ that maps a text prompt $c$ to a generated video $v$. Our goal is to fine-tune $\pi_\theta$ to maximize the expected reward $R(v)$ while staying close to the base reference policy $\pi_{\mathrm{ref}}$ via a KL penalty:
\begin{equation}
\label{eq:rl-objective}
\max_\theta \; \mathbb{E}_{c \sim \mathcal{D},\, v \sim \pi_\theta(\cdot \mid c)} \big[ R(v) \big] - \lambda \, \mathrm{KL}\!\big(\pi_\theta \,\|\, \pi_{\mathrm{ref}}\big),
\end{equation}
where $\lambda$ controls the strength of the KL regularization to prevent mode collapse and preserve general video generation capability. The prompt distribution $\mathcal{D}$ is a curated set of human-motion-specific prompts covering diverse actions and scenarios (details in \Cref{sec:rl_results}). 

\noindent{\textbf{Policy optimization.}}
We adopt the forward-process RL formulation of DiffusionNFT~\citep{zheng2025diffusionnft}, following Astrolabe~\citep{zhang2026astrolabe}. Given a generated video $v_0 \sim \pi_\theta(\cdot \mid c)$ with normalized reward $\tilde{r} \in [0,1]$, a noisy version $v^t$ is constructed at a randomly sampled timestep $t$. Using the current velocity predictor $v_\theta$ and the old predictor $v_{\theta_{\text{old}}}$, implicit positive and negative policies are defined via interpolation:
\begin{equation}
\label{eq:interpolation}
v^+ = (1-\beta)\, v_{\theta_{\text{old}}} + \beta\, v_\theta, \quad v^- = (1+\beta)\, v_{\theta_{\text{old}}} - \beta\, v_\theta,
\end{equation}
where $\beta$ controls the interpolation strength. The policy loss contrasts these implicit policies against the target forward velocity $v_{\text{target}}$:
\begin{equation}
\label{eq:policy-loss}
\mathcal{L}_{\text{policy}} = \tilde{r}\, \|v^+ - v_{\text{target}}\|_2^2 + (1 - \tilde{r})\, \|v^- - v_{\text{target}}\|_2^2.
\end{equation}
This trajectory-free formulation requires only clean generated samples and avoids backpropagating through the reverse sampling chain, enabling memory-efficient and solver-agnostic training.

\noindent{\textbf{Reward integration.}}
The reward $R_{\mathrm{motion}}(v)$ in \Cref{eq:rl-objective} follows the
equally weighted aggregation defined in \Cref{eq:reward-agg}.
Because the reward is decomposed into three independent axes, we can diagnose per-dimension improvement during training and detect reward hacking early. For example, if one axis improves disproportionately at the expense of others, the decomposition makes this immediately visible.

\section{Experiments}
\label{sec:experiments}

We organize our experiments around two central questions. First, we evaluate
whether our physics-grounded 3D metrics better align with human judgments of
synthetic human video quality than existing video metrics (\Cref{sec:human_alignment}). Second, we use the
same metrics as rewards for RL post-training and evaluate whether they improve
human motion video generation (\Cref{sec:rl_results}).

\begin{figure}[t]
    \centering
    \includegraphics[width=\linewidth]{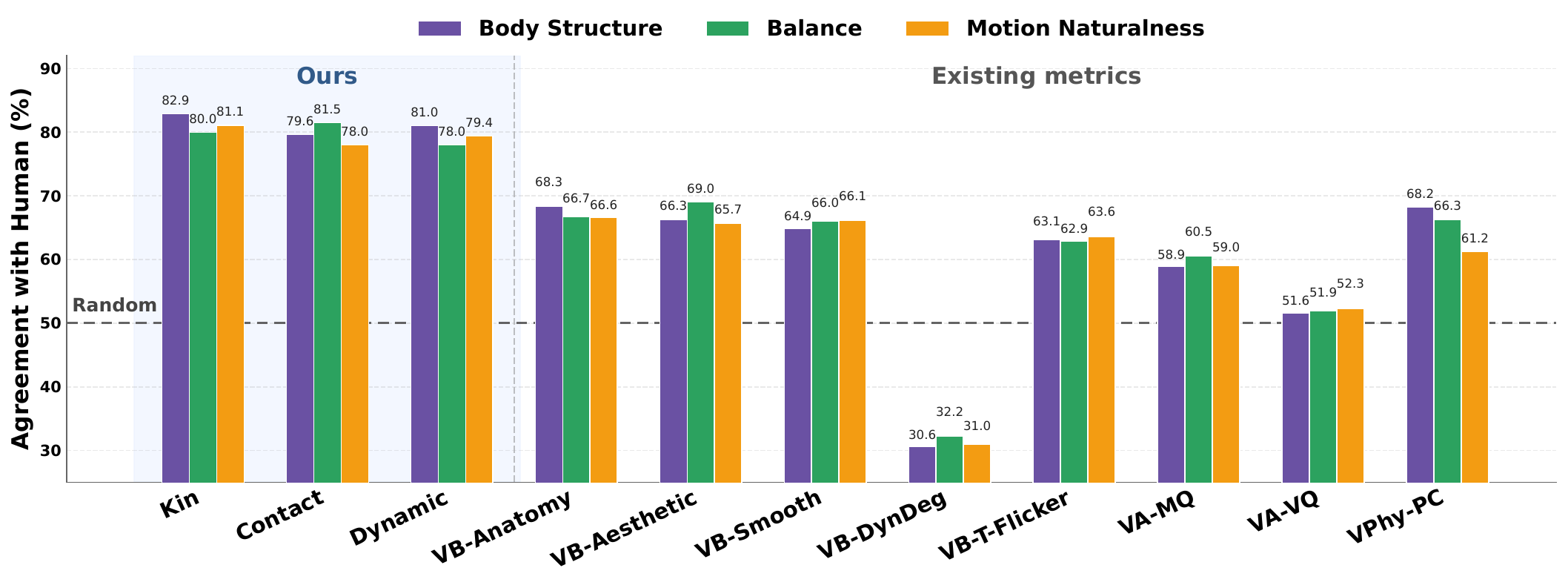}
    \caption{
    \textbf{Agreement with human judgments for motion quality.}
    Our metrics achieve the highest agreement compared with perceptual (VBench / VBench2) and learned (VideoAlign, VideoPhy) video metrics across three human-evaluation questions: body structure, balance, and motion naturalness.  
    }
    \vspace{-3mm}
    \label{fig:human_alignment}
\end{figure}

\subsection{Human Alignment Results of \methodName}
\label{sec:human_alignment}
\noindent\textbf{Experiment setup.}
We compare against existing evaluators, including
VBench~\citep{huang2024vbench}, VBench-2.0~\citep{zheng2025vbench2},
VideoAlign~\citep{liu2025videoalign}, and
VideoPhy~\citep{bansal2024videophy}.
 To construct the annotation set, we sample prompts
from the Motion-X dataset~\citep{lin2023motionx} and generate videos under the
same prompt using six baseline video models: Wan-2.1 1.3B, Wan-2.2 5B,
Wan-2.2 14B~\citep{wan2025wan}, Causal Forcing-1.3B~\citep{zhu2026causal},
EchoMotion-5B~\citep{yang2025echomotion}, and
FastWan~\citep{zhang2025fast}. For each sample, we form video pairs from
different models based on the same prompt, so that annotators compare motion
quality under identical text conditions rather than across different
conditions. From these candidate pairs, we randomly select $1{,}200$
comparisons for annotation. Each annotation presents two anonymized videos
side by side with randomized left/right order, together with the shared text
prompt. We recruit six annotators and ask them to judge which video is better
along three criteria (body structure, balance, and motion naturalness) corresponding to the failure modes in \Cref{sec:failure}.
For each
criterion, annotators choose among ``Video A'', ``Video B'', or ``Tie'',
allowing ambiguous cases to be excluded from decisive metric-alignment
comparisons.
Additional annotation details are provided in Appendix~\ref{app:human_detail}.

\begin{table*}[t]
\centering
\scriptsize
\setlength{\tabcolsep}{3.0pt}
\renewcommand{\arraystretch}{1.08}
\caption{
\textbf{Correlation with human judgments for motion quality.}
We report Spearman's rank correlation ($\rho$) between automatic metric scores
and human judgments across three questions. Best
results are in \textbf{bold}, and second-best results are \underline{underlined}.
}
\label{tab:human_alignment}
\resizebox{\textwidth}{!}{
\begin{tabular}{lcccccccccccc}
\toprule
{Question}
& \multicolumn{5}{c}{\textit{VBench/VBench2}}
& \multicolumn{2}{c}{\textit{VideoAlign}}
& \multicolumn{1}{c}{\textit{VideoPhy}}
& \multicolumn{4}{c}{\textit{\methodName{} (Ours)}} \\
\cmidrule(lr){2-6}
\cmidrule(lr){7-8}
\cmidrule(lr){9-9}
\cmidrule(lr){10-13}
& Aes.~Q. & Mot.~Sm. & Temp.~Fl. & Dyn.~Deg. & Hum. An.
& VA-MQ & VA-VQ
& VP-PC
& Kin. & Con. & Dyn. & Overall \\

\midrule
Body Structure
& $+0.248$
& $+0.236$
& $+0.175$
& $-0.135$
& $+0.273$
& $+0.155$
& $+0.045$
& $+0.168$
& $\underline{+0.369}$
& $+0.290$
& $+0.367$
& $\mathbf{+0.391}$ \\
Balance
& $+0.238$
& $+0.246$
& $+0.171$
& $-0.118$
& $+0.235$
& $+0.175$
& $+0.053$
& $+0.142$
& $+0.314$
& $\mathbf{+0.337}$
& $+0.316$
& $\underline{+0.333}$ \\
Motion Naturalness
& $+0.260$
& $+0.274$
& $+0.208$
& $-0.149$
& $+0.278$
& $+0.156$
& $+0.067$
& $+0.108$
& $+0.375$
& $+0.281$
& $\underline{+0.389}$
& $\mathbf{+0.402}$ \\
All Questions
& $+0.248$
& $+0.252$
& $+0.185$
& $-0.135$
& $+0.262$
& $+0.161$
& $+0.055$
& $+0.138$
& $+0.353$
& $+0.290$
& $\underline{+0.358}$
& $\mathbf{+0.376}$ \\
\bottomrule
\end{tabular}
}
\end{table*}

\noindent\textbf{Quantitative results.} As shown in \Cref{fig:human_alignment} and \Cref{tab:human_alignment}, our
physics-grounded metrics achieve the strongest and most consistent alignment
with human judgments under both pairwise agreement and Spearman's rank
correlation ($\rho$).
In the agreement results, kinematic feasibility performs best on body structure
($82.9\%$), while contact and dynamic feasibility remain consistently high
across all criteria ($78$--$81\%$); in contrast, existing perceptual and learned
metrics mostly remain in the $50$--$66\%$ range, with Dynamic Degree near chance.
The same trend holds in correlation: overall feasibility ($R_{\mathrm{motion}}$) achieves the highest
aggregate correlation ($+0.376$), outperforming the strongest existing metric,
VBench2 Human Anatomy ($+0.262$), as well as VideoAlign-MQ ($+0.161$) and
VideoPhy-PC ($+0.138$). The individual components also align with their intended
diagnostic roles: kinematic feasibility is strongest for body-structure
judgments, contact feasibility is strongest for balance, and dynamic feasibility
is highly correlated with motion naturalness, providing non-redundant diagnostic signals that serve as stable, interpretable rewards for RL-based post-training (\Cref{sec:rl_results}). \change{We further evaluate the robustness of PhyMotion to severe anatomical distortions and 3D reconstruction errors (Appendix~\ref{app:reconstruction_robustness}), as well as its extensibility to non-flat terrain (Appendix~\ref{app:terrain_robustness}).}

\subsection{RL Post-Training Results of \methodName}
\label{sec:rl_results}
\subsubsection{Experiment Setup}
We apply our structured 3D motion reward to two
distilled 4-step Wan 1.3B backbones: FastWan~\citep{zhang2025fast}
(bidirectional) and Causal Forcing~\citep{zhu2026causal} (autoregressive). We
recover SMPL motion with GVHMR~\citep{shen2024gvhmr} and compute physical
feedback in MuJoCo~\citep{todorov2012mujoco}. We fine-tune with LoRA
adapters~\citep{hu2022lora} using RL post-training; full hyperparameters are
provided in Appendix~\ref{app:training_detail}. Training uses $8$ NVIDIA A100
80\,GB GPUs for $330$ steps, taking $2.5$ days.
We compare against EchoMotion~\citep{yang2025echomotion}, Causal
Forcing~1.3B~\citep{zhu2026causal}, FastWan~1.3B~\citep{zhang2025fast}, and
Wan models including Wan~2.1~1.3B, Wan~2.2~5B, and
Wan~2.2~14B~\citep{wan2025wan}. To compare against VLM-based reward optimization, we also fine-tune the Causal Forcing 1.3B model with the VideoAlign Motion Quality (MQ) reward under the same training setting. We evaluate on two prompt sets: a human-motion prompt set of $22{,}471$ prompts
derived from Motion-X~\citep{lin2023motionx}, and the standard VBench and
VBench-2.0 prompt sets~\citep{huang2024vbench,zheng2025vbench2}. We report three metric groups:
VBench/VBench-2.0 metrics, VLM-based rewards
(VideoAlign-MQ/VQ~\citep{liu2025videoalign} and
VideoPhy-PC~\citep{bansal2024videophy}), and our \methodName{} from \Cref{sec:Physics-Grounded Evaluation}.

\begin{table*}[t]
\centering
\caption{
\textbf{Main results on human-motion video generation.}
We compare open-source baselines, a VLM-reward-trained baseline, and our structured 3D reward training. We report VBench perceptual metrics, VideoAlign scores, VideoPhy Physical Commonsense, and our structured 3D reward scores. All metrics are higher-is-better. \textbf{Bold}: best; \underline{underline}: second best. 
}
\label{tab:main}
\resizebox{\textwidth}{!}{%
\begin{tabular}{@{}l ccccc cc c cccc@{}}
\toprule
\multirow{2}{*}{Method}
& \multicolumn{5}{c}{\textit{VBench/VBench2}}
& \multicolumn{2}{c}{\textit{VideoAlign}}
& \multicolumn{1}{c}{\textit{VideoPhy}}
& \multicolumn{4}{c}{\textit{\methodName{} (Ours)}} \\
\cmidrule(lr){2-6}
\cmidrule(lr){7-8}
\cmidrule(lr){9-9}
\cmidrule(lr){10-13}
& Mot.~Sm. & Dyn.~Deg. & Aes.~Q. & Hum.~An. & Temp.~Fl.
& VA-MQ & VA-VQ
& VP-PC
& Kin. & Con. & Dyn. & Overall \\
\midrule

\sectionrow{Open-source baselines}

{Wan2.2 14B~\citep{wan2025wan}}
& {0.9892} & {0.393} & {0.5141} & {\underline{0.943}} & {{0.9857}}
& {1.133} & {0.557}
& {3.97}
& {0.881} & {0.733} & {0.912} & {0.842} \\

{Wan2.2 5B~\citep{wan2025wan}}
& {0.9846} & {0.598} & {0.4622} & {0.823} & {0.9803}
& {1.069} & {0.514}
& {3.91}
& {0.880} & {0.710} & {0.913} & {0.834} \\

{EchoMotion 5B~\citep{yang2025echomotion}}
& {0.9850} & {{0.683}} & {0.4756} & {0.836} & {0.9777}
& {1.044} & {0.565}
& {3.74}
& {0.888} & {0.702} & {0.918} & {0.836} \\

Wan 1.3B~\citep{wan2025wan}
& 0.9868 & \underline{0.700} & 0.4758 & 0.834 & 0.9787
& 1.213 & {0.621}
& 3.86
& 0.862 & 0.705 & 0.904 & 0.824 \\

Causal Forcing 1.3B~\citep{zhu2026causal}
& {0.9895} & 0.631 & \underline{0.5391} & 0.886 & 0.9819
& {1.241} & 0.575
& {4.06}
& 0.927 & {0.739} & {0.948} & {0.871} \\

FastWan 1.3B~\citep{zhang2025fast}
& 0.9847 & \textbf{0.928} & 0.5035 & 0.876 & 0.970
& 1.307 & 0.680
& 3.90
& 0.908 & 0.734 & 0.916 & 0.853 \\

\midrule

\sectionrow{VLM-reward trained baseline} 

VideoAlign-MQ~\citep{liu2025videoalign}
& 0.9904 & 0.570 & 0.5229 & 0.910 & 0.9835
& \textbf{1.549} & \textbf{0.912}
& 4.12
& 0.916 & 0.753 & 0.946 & 0.878 \\

\midrule

\sectionrow{\methodName{} (Ours) trained with}

Causal Forcing 1.3B
& \makecell{\textbf{0.9956}\\[-1pt]\scriptsize(+0.6\%)}
& \makecell{0.411\\[-1pt]\scriptsize(-34.9\%)}
& \makecell{\textbf{0.5589}\\[-1pt]\scriptsize(+3.7\%)}
& \makecell{\textbf{0.951}\\[-1pt]\scriptsize(+7.3\%)}
& \makecell{\textbf{0.9957}\\[-1pt]\scriptsize(+1.4\%)}
& \makecell{\underline{1.313}\\[-1pt]\scriptsize(+5.8\%)}
& \makecell{\underline{0.720}\\[-1pt]\scriptsize(+25.2\%)}
& \makecell{\underline{4.29}\\[-1pt]\scriptsize(+5.7\%)}
& \makecell{\textbf{0.982}\\[-1pt]\scriptsize(+5.9\%)}
& \makecell{\textbf{0.809}\\[-1pt]\scriptsize(+9.4\%)}
& \makecell{\textbf{0.988}\\[-1pt]\scriptsize(+4.2\%)}
& \makecell{\underline{0.902}\\[-1pt]\scriptsize(+3.5\%)} \\
FastWan 1.3B
& \makecell{\underline{0.9927}\\[-1pt]\scriptsize(+0.8\%)}
& \makecell{0.420\\[-1pt]\scriptsize(-54.7\%)}
& \makecell{0.5218\\[-1pt]\scriptsize(+3.6\%)}
& \makecell{0.867\\[-1pt]\scriptsize(-1.0\%)}
& \makecell{\underline{0.9928}\\[-1pt]\scriptsize(+2.4\%)}
& \makecell{{1.318}\\[-1pt]\scriptsize(+0.8\%)}
& \makecell{0.687\\[-1pt]\scriptsize(+1.0\%)}
& \makecell{\textbf{4.39}\\[-1pt]\scriptsize(+12.6\%)}
& \makecell{\textbf{0.982}\\[-1pt]\scriptsize(+8.1\%)}
& \makecell{\underline{0.805}\\[-1pt]\scriptsize(+9.7\%)}
& \makecell{\underline{0.952}\\[-1pt]\scriptsize(+3.9\%)}
& \makecell{\textbf{0.913}\\[-1pt]\scriptsize(+7.0\%)} \\

\bottomrule
\end{tabular}%
}
\vspace{-3mm}
\end{table*}

\subsubsection{Quantitative and Qualitative Results}
\noindent\textbf{Quantitative results on automatic metrics.}
\Cref{tab:main} shows that our structured 3D reward improves human-motion
quality beyond the metrics used for optimization. Compared with the Causal
Forcing 1.3B base model, our RL-post-trained model improves all external metrics
except VBench Dynamic Degree, including VBench motion smoothness, aesthetic
quality, human action, temporal flickering, VideoAlign, and VideoPhy. Notably,
it improves VideoAlign video quality by $+25.2\%$ and VideoPhy physical
commonsense by $+5.7\%$, and even outperforms larger 5B/14B reference models on
most metrics. 
\begin{wrapfigure}{r}{0.5\textwidth}
    \centering
     \vspace{-2mm}
\includegraphics[width=\linewidth]{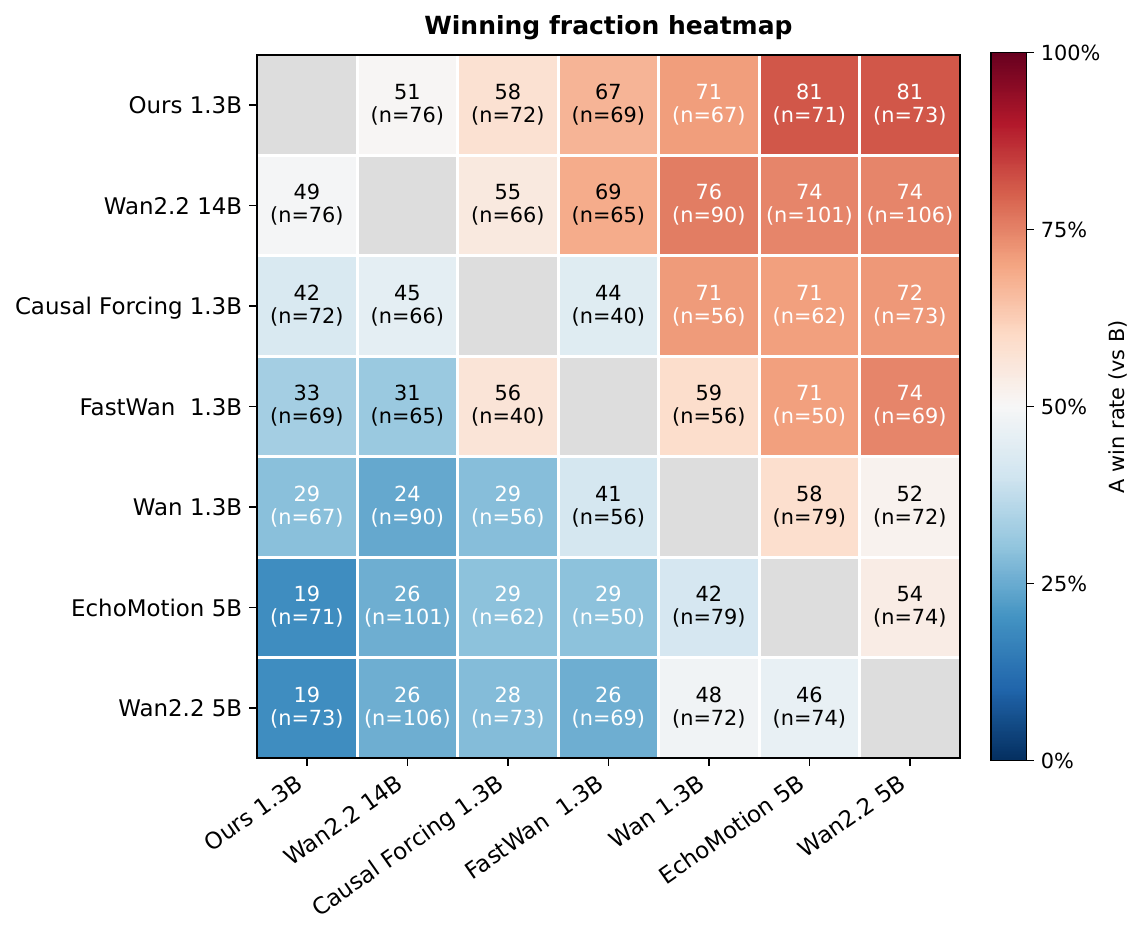}
    \vspace{-2mm}
    \caption{
    \textbf{Winning fraction heatmap of different models.}
    Each cell shows the win rate between models on the overall
    human preference evaluation, with ties counted as $0.5$. 
     \vspace{-3mm}
    }
    \label{fig:human_preference_heatmap}
\end{wrapfigure} 
As expected, these gains are also reflected in our \methodName{}
rewards, with improvements of $+3.5\%$ in overall
feasibility. 
The same trend also
holds for FastWan 1.3B: after post-training with \methodName{}, the model improves
motion smoothness, aesthetic quality, temporal flickering, VideoAlign, VideoPhy,
and all \methodName{} scores, including a $+7.0\%$ gain in overall
feasibility. \change{The decrease in VBench Dynamic Degree may reflect both reduced motion artifacts and more conservative motion. To investigate this trade-off, we evaluate instruction execution using Gemini and motion pacing through human judgments (Appendix~\ref{app:motion_suppression}). Gemini-based evaluation shows that full action execution improves from 13.9\% to 21.5\% after RL, while human judgments indicate that most videos remain appropriately paced (88\% vs. 87\%), despite a modest increase in overly slow motions (3\% to 8\%).} Furthermore, compared
with the VLM-reward-trained VideoAlign-MQ baseline, our method achieves higher
VideoPhy physical commonsense and stronger structured 3D feasibility across all
dimensions, suggesting that explicit 3D physical rewards provide a more targeted
training signal than generic VLM-based video rewards.

\noindent\textbf{Quantitative results on human preference evaluation.}
To verify that improvements in automatic metrics translate to human-perceived
motion quality, we further conduct a human preference evaluation across
representative baseline models. We randomly sample $1{,}487$ video pairs and ask
human annotators the same preference questions as in~\Cref{sec:human_alignment}.
As shown in~\Cref{tab:human_preference}, our RL-post-trained model achieves the
highest overall human preference Elo score, outperforming all baselines,
including the much larger Wan2.2 14B model.

Beyond aggregate Elo rankings, we additionally compute pairwise human preference win rates between models. For each model pair, we count a win as $1$, a loss as $0$, and a tie as $0.5$, and report the average win rate between models. As shown in~\Cref{fig:human_preference_heatmap}, our \methodName{} optimized model consistently wins against all compared baselines in direct head-to-head comparisons, including the much larger Wan2.2 14B model. This indicates that the
human preference improvement of \methodName{} is not driven by a single favorable matchup, but is broadly consistent across diverse baseline comparisons.

\noindent\textbf{Qualitative comparison.}
\Cref{fig:qualitative} provides qualitative comparisons across diverse
human-motion prompts, including (a) martial arts, (b) multi-human dancing, (c) soccer, (d) handstand, (e) figure skating, and (f) floor
exercise / side lifting. Compared with baseline models, our model better
preserves the intended action structure over time. 
In challenging motions
such as handstand, figure skating, and floor exercises
(\Cref{fig:qualitative}d--f), baseline models often exhibit unstable body
support, floating feet or hands, implausible limb bending, or sudden pose
discontinuities. In contrast, our model more consistently maintains physically plausible motions.

\begin{table}[t]
\vspace{-8pt}
\centering

\begin{minipage}[t]{0.44\textwidth}

\centering
\scriptsize
\setlength{\tabcolsep}{4pt}
\renewcommand{\arraystretch}{1.05}
\caption{
\textbf{Human preference evaluation.}
We report Elo ratings computed from pairwise human preferences. Higher is better.
}
\label{tab:human_preference}\begin{tabular}{lcccc}
\toprule
\textbf{Model} & \textbf{Body} & \textbf{Bal.} & \textbf{Motion} & \textbf{All Q} \\
\midrule
Wan2.2 5B & $1376$ & $1388$ & $1384$ & $1383$ \\
EchoMotion 5B & $1386$ & $1403$ & $1374$ & $1387$ \\
Wan 1.3B & $1429$ & $1440$ & $1411$ & $1427$ \\
FastWan 1.3B & $1526$ & $1521$ & $1528$ & $1525$ \\
Causal 1.3B & $1562$ & $1546$ & $1553$ & $1553$ \\
Wan2.2 14B & $1600$ & $1593$ & $1618$ & $1604$ \\
\midrule
Ours 1.3B & $\mathbf{1620}$ & $\mathbf{1610}$ & $\mathbf{1632}$ & $\mathbf{1621}$ \\
\bottomrule
\end{tabular}
\end{minipage}
\hfill
\begin{minipage}[t]{0.53\textwidth}

\centering
\caption{
\textbf{Ablation results on rewards.} \textbf{VB}, \textbf{VA}, and \textbf{VP} denote compact summary metrics for VBench average, VideoAlign average, and VideoPhy Physical Commonsense, respectively.
}
\label{tab:ablation_compact}
\footnotesize
\setlength{\tabcolsep}{3.0pt}
\renewcommand{\arraystretch}{1.03}
\begin{tabular}{lccccccc}
\toprule
Reward & VB & VA & VP & Kin. & Con. & Dyn. & Ours \\
\midrule
Average
& \textbf{0.782} & \textbf{0.979} & \textbf{4.12}
& \underline{0.954} & \underline{0.763} & {0.943}  & \textbf{0.883} \\
Only Kin.
& 0.767 & 0.904 & 3.86
& \textbf{0.963} & {0.730} & \underline{0.951}  & \underline{0.881} \\
Only Con.
& \underline{0.773} & 0.897 & 3.74
& 0.911  & \textbf{0.772}  & 0.917 & 0.867 \\
Only Dyn.
& 0.766 & \underline{0.915} & \underline{3.91}
& 0.948 & 0.734 &  \textbf{0.958}  & {0.880} \\
\bottomrule
\end{tabular}
\end{minipage}

\vspace{-8pt}
\end{table}

\begin{figure}[t]
    \centering
    \includegraphics[width=\linewidth]{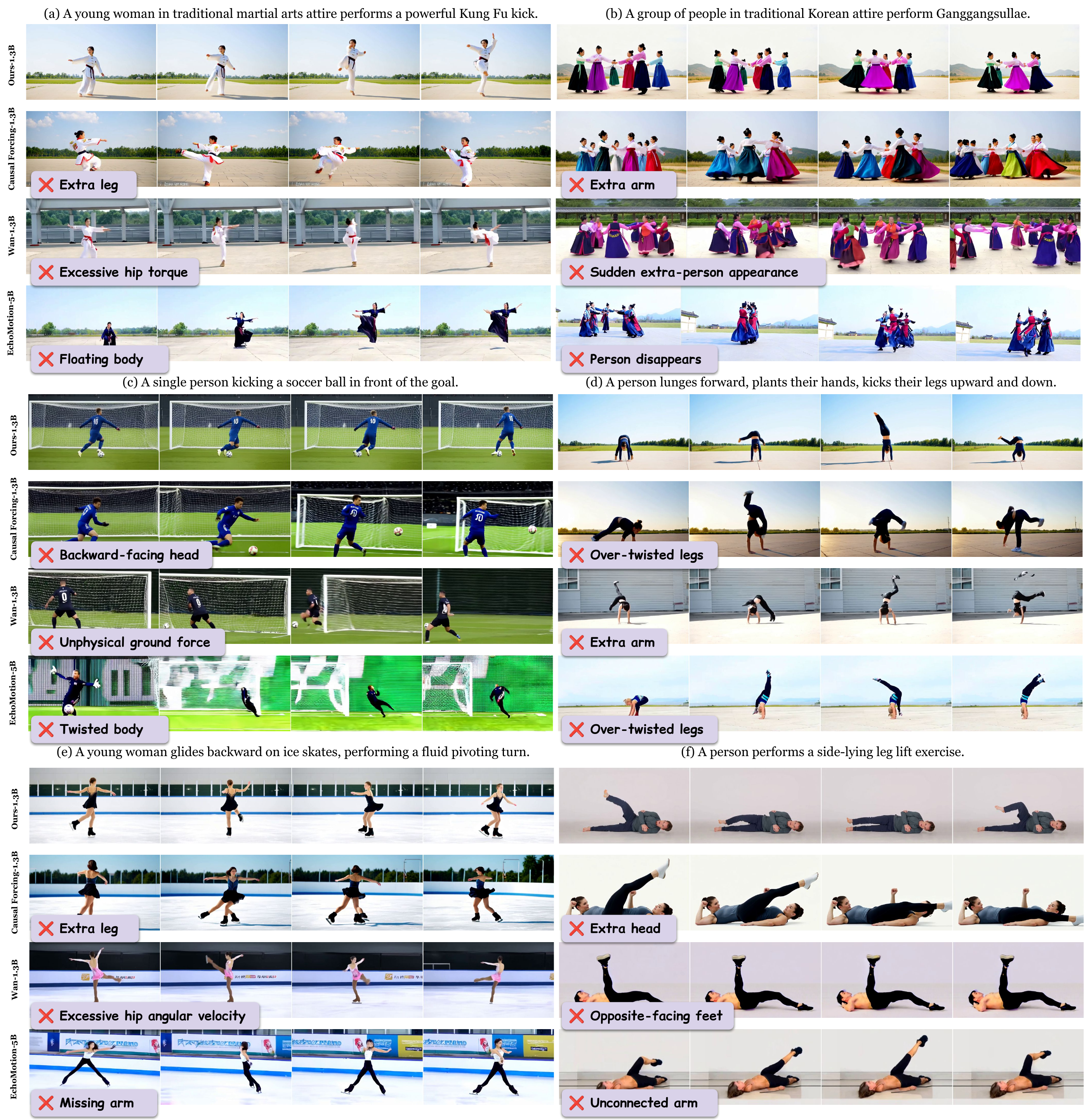}
    \caption{
\textbf{Qualitative comparison on diverse human-motion prompts.} We compare our model against baseline models across challenging motion
categories. Baseline models often produce artifacts, while our model maintains more physically plausible motion.
}   
    \label{fig:qualitative}
\vspace{-10pt}
\end{figure}

\subsubsection{Additional Analysis}
\label{sec:additional_analysis}

We further analyze \methodName{} from three perspectives: the contribution of
each reward component,  whether reward
post-training preserves general-domain video generation ability, and
the computational overhead of our reward during training.

\begin{wraptable}{r}{5.5cm}
\centering
\caption{
\textbf{Generalization to general prompts.}  Gray rows denote reference models at larger scales.
}
\label{tab:general_vbench}
\footnotesize
\setlength{\tabcolsep}{4pt}
\renewcommand{\arraystretch}{1.03}
\begin{tabular}{lcc}
\toprule
Model & VBench & VBench-2.0 \\
\midrule
\gray{Wan 2.2 5B}
& \gray{0.607} & \gray{0.403} \\
\gray{Wan 2.2 14B}
& \gray{0.657} & \gray{0.510} \\
\midrule
Wan 1.3B
& 0.630 & 0.456 \\
Causal Forcing 1.3B
& \textbf{0.657} & \underline{0.458} \\
Ours
& \underline{0.654} & \textbf{0.462} \\
\bottomrule
\end{tabular}
\end{wraptable}
\noindent\textbf{Effect of individual reward components.}
We ablate each reward component by training single-reward variants from the same
Causal Forcing base model, using identical hyperparameters and 120 training
steps for all variants. As shown in \Cref{tab:ablation_compact}, each
single-reward variant primarily improves its corresponding feasibility
dimension. However, optimizing any single component alone does not yield the best
overall performance. In contrast, the full reward achieves the highest overall
feasibility while remaining competitive across all individual dimensions,
indicating that jointly optimizing kinematic, contact, and dynamic feasibility
produces a better-balanced human-motion prior.

\change{
\noindent\textbf{Generalization to broader prompts.}
To assess whether our human-centric reward compromises general video
generation quality, we evaluate on standard VBench and VBench-2.0
prompts. As shown in \Cref{tab:general_vbench}, \methodName{}
maintains comparable VBench performance to the Causal Forcing 1.3B
backbone ($0.654$ vs.\ $0.657$) while slightly improving VBench-2.0
($0.462$ vs.\ $0.458$). These results suggest that optimizing
physical plausibility does not substantially sacrifice general
generation capabilities. Per-category results are provided in
Appendix~\ref{app:per_category_vbench}.

\begin{table}[t]
\centering
\caption{
\textbf{Reward computation overhead.}
We report the per-video reward computation time and the effective training overhead introduced by each reward module. 
}
\label{tab:reward_overhead}
\small
\setlength{\tabcolsep}{5pt}
\renewcommand{\arraystretch}{1.05}
\begin{tabular}{lcccc}
\toprule                                                                                    
   & HPSv3 & VideoAlign-MQ & VideoPhy-PC & \methodName{} (Ours) \\
  \midrule                                                                                                                            
  Time / video (s)   & 4.72 & 0.25 & 0.20 & 2.80 \\                                    
  Effective overhead & 35\%  & 0.2\%  & 0.2\%  & 7\%  \\
  \bottomrule           
\end{tabular}
\vspace{-3mm}
\end{table}

\noindent\textbf{Reward computation overhead.}
As shown in \Cref{tab:reward_overhead}, \methodName{} requires
$2.80$s per video but introduces only $7\%$ effective training
overhead through pipelined reward computation, compared with
$35\%$ for HPSv3.
Further details are provided in Appendix~\ref{app:reward_overhead_details}.
}

\section{Conclusion}
\label{sec:conclusion}
We introduce \methodName, a structured, fine-grained motion reward for physics-grounded human video generation. Instead of relying only on perceptual or learned video-level
signals, \methodName{} lifts generated videos into 3D human motion, grounds them in a
physics simulator, and evaluates motion feasibility through kinematic,
contact/balance, and dynamic consistency. Experiments show that \methodName{} aligns better with human judgments than existing video rewards, and our post-training experiment also shows that \methodName{} serves as an effective RL post-training objective, improving automatic metrics performance and human preference scores while preserving general video quality across different video-generation backbones. 

\begin{ack}
This work was supported by ONR Grant N00014-23-1-
2356, ARO Award W911NF2110220, and NSF-AI Engage Institute DRL-2112635. The views contained in this article are those of the authors and not of the funding agency.
\end{ack}

\bibliographystyle{plainnat}
\bibliography{ref}

\appendix

\clearpage
\section*{Appendix}

\section{Limitations}
\label{sec:limitations}

This work studies structured reward design for improving human-motion realism
in video generation. As with other reward-based post-training methods, our
approach may be susceptible to reward hacking. Although our quantitative
analysis shows improved action execution and largely preserved motion pacing
after RL post-training (Appendix~\ref{app:motion_suppression}), isolated
failure cases still exhibit motion suppression or simplification of challenging
actions (Fig.~\ref{fig:limitations}(a)). Moreover, our reward focuses on
human-body motion without explicitly modeling human--object interactions,
which may lead to physically inconsistent object manipulation
(Fig.~\ref{fig:limitations}(b)). Finally, our post-training primarily
covers motions on flat surfaces, limiting generalization to out-of-distribution
scenarios such as stair climbing (Fig.~\ref{fig:limitations}(c)).
Although our evaluation framework can incorporate non-flat terrain geometry,
such scenarios remain underrepresented during training. Future work could
address these limitations through more diverse training data and reward
signals that explicitly account for action completion, object interactions,
and complex environments.

\section{Broader Impact}
\label{app:boarder_impact}

While \methodName{} aims to improve the physical realism of human motion in video
generation, it is important to consider its broader ethical, legal, and societal
implications.

\paragraph{Potential benefits.}
Improving the physical realism of human motion in generated videos can benefit
applications that rely on reliable human-centric simulation, including
education, accessibility, sports and exercise instruction, entertainment, virtual
communication, and embodied-agent training. In addition, because \methodName{}
decomposes motion quality into kinematic, contact, and dynamic feasibility, it
provides more interpretable feedback than a single holistic video score. This
can help researchers and practitioners diagnose specific motion failures,
develop safer generation systems, and better understand when synthetic human
videos are physically plausible or unreliable.

\paragraph{Deceptive misuse.}
More realistic human motion may make generated videos appear more convincing,
which could increase the risk of misuse in misinformation, impersonation, or
fraudulent media. Although our work focuses on evaluating and improving physical
plausibility rather than identity preservation or face realism, the resulting
models may still contribute to more realistic synthetic human videos. We therefore
encourage responsible release practices, clear usage guidelines, watermarking or
provenance mechanisms when appropriate, and safeguards against deceptive or
non-consensual use.

\paragraph{Consent, privacy, and human likeness.}
Human-centric video generation raises concerns around consent and privacy,
especially when generated videos resemble real individuals or depict sensitive
activities. Our experiments use text prompts and generated videos for evaluation,
rather than collecting private human video data for model training. Nevertheless,
future applications of human-motion video generation should respect individual
consent, avoid unauthorized likeness generation, and follow applicable privacy
and publicity-rights regulations.

\paragraph{Bias and fairness.}
\methodName{} builds on existing video generation backbones and 3D human
reconstruction tools, which may inherit biases from their training data and
model design. These biases may affect motion quality across different body
types, clothing styles, activities, camera viewpoints, or cultural contexts.
Although our reward is designed to evaluate physical feasibility in 3D, it does
not by itself guarantee fairness across all populations or motion categories.
Careful evaluation across diverse human appearances, body shapes, and actions is
needed before deployment.

\begin{figure}[t]
    \centering
    \includegraphics[width=\linewidth]{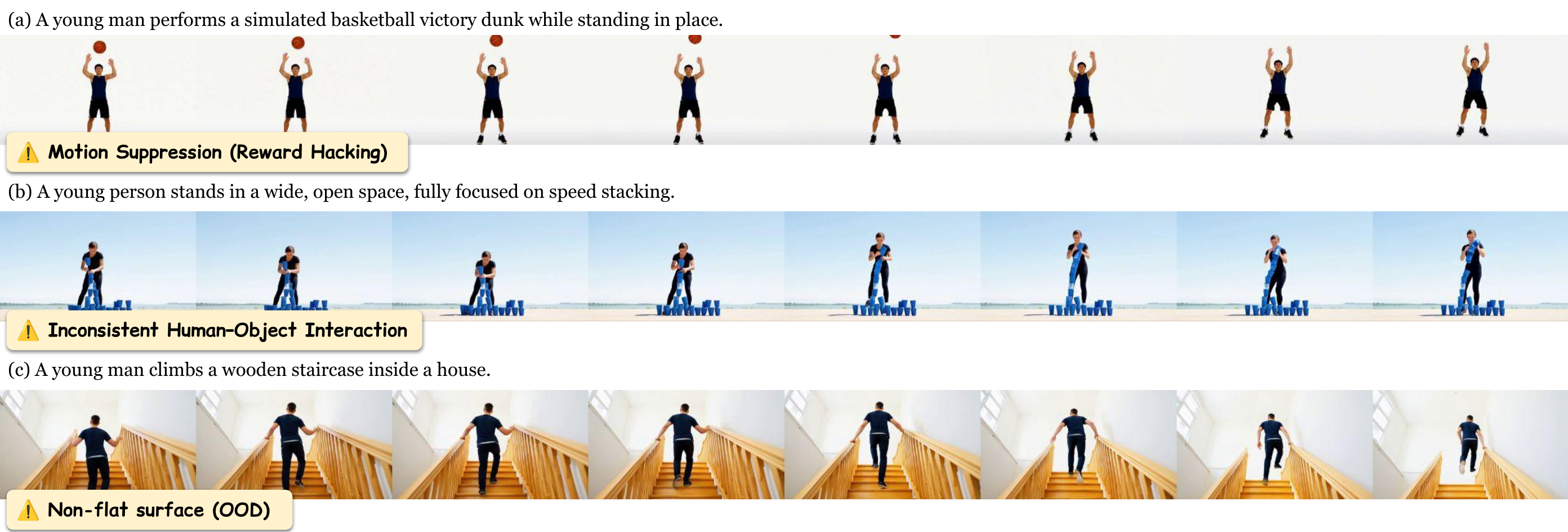}
    \caption{\textbf{Qualitative limitations of PhyMotion.}
    Representative failure cases involving
    (a) motion suppression due to reward hacking,
    (b) physically inconsistent human--object interactions,
    and (c) out-of-distribution non-flat terrain.}
    \label{fig:limitations}
\end{figure}

\section{Additional Qualitative Results}
\subsection{Qualitative Analysis of Individual Reward Components}
\label{app:qualitative_submetrics}

We further provide qualitative examples to illustrate the diagnostic role of each
\methodName{} submetric. As shown in \Cref{fig:qualitative_by_metric}, each
example is selected by filtering for videos that score poorly on one target
submetric while scoring relatively well on the other two. This isolates failure
cases primarily captured by a single component.

The examples show that the three submetrics capture complementary artifacts.
Kinematic feasibility~\Cref{fig:qualitative_by_metric}(a) detects articulated-body errors such as self-penetration,
extra limbs and impossible joint angles. Contact/balance feasibility~\Cref{fig:qualitative_by_metric}(b) captures unstable support, balance
violations, and foot sliding. Dynamic feasibility~\Cref{fig:qualitative_by_metric}(c)  identifies motions requiring
implausible forces and torques, such as excessive horizontal ground force. These results
support the need for a decomposed reward: each component provides an interpretable
signal for a distinct class of physical failures.

\begin{figure}[t]
    \centering
    \includegraphics[width=0.9\linewidth]{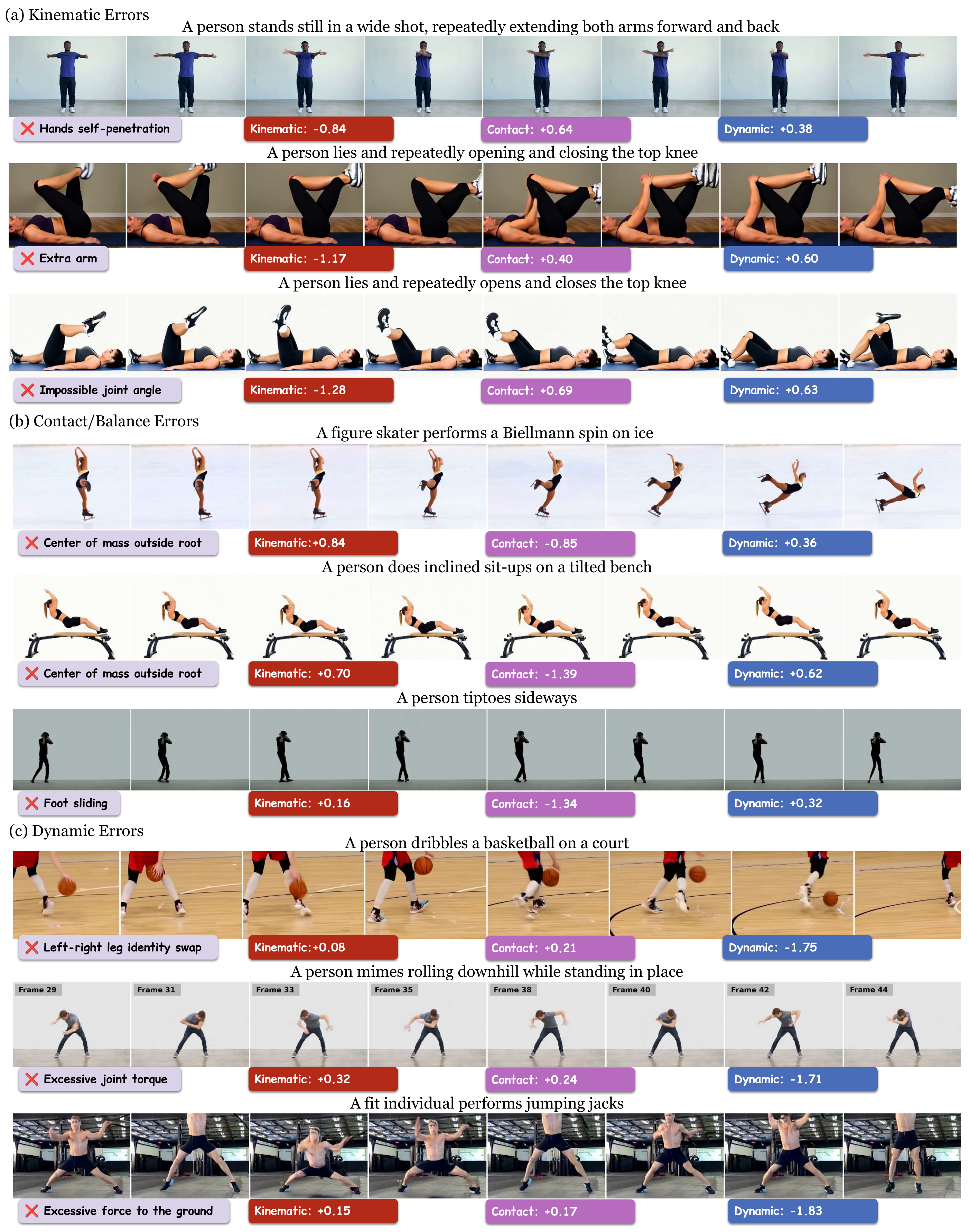}
    \caption{
    \textbf{Qualitative examples selected by individual \methodName{} submetrics.}
    Each row shows a generated video that scores poorly on one target submetric
    while scoring relatively well on the other two. The examples show
    complementary failures captured by kinematic feasibility, contact/balance
    feasibility, and dynamic feasibility. All scores are reported in z-scores after normalization.
    }
    \label{fig:qualitative_by_metric}
\end{figure}

\section{Per-category Comparison on General Video Benchmarks.}
\label{app:per_category_vbench}
\begin{figure*}[t]
\centering
\begin{subfigure}[t]{0.48\textwidth}
    \centering
    \includegraphics[width=0.95\linewidth]{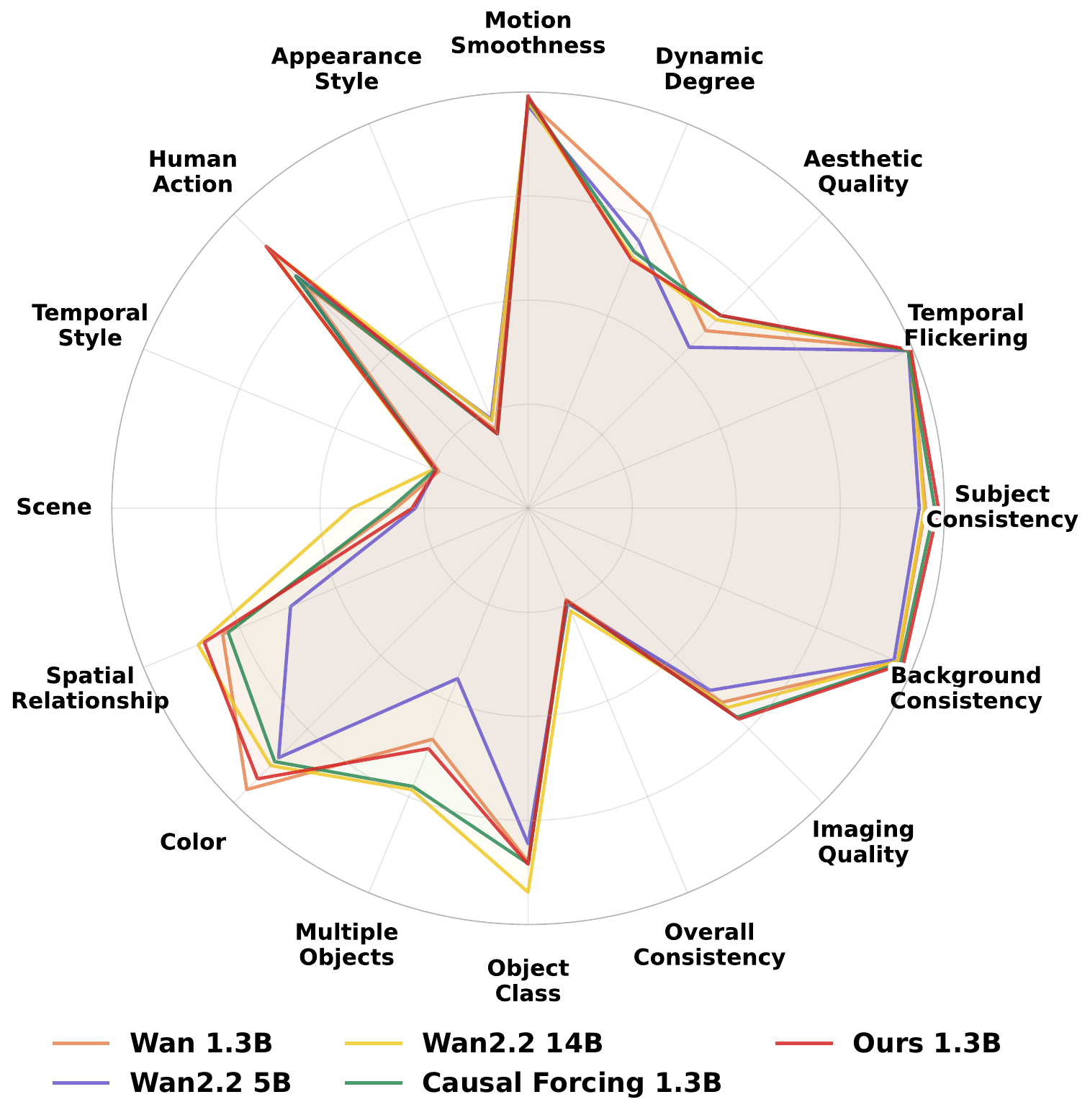}
    \caption{VBench per-dimension comparison.}
    \label{fig:vbench_radar}
\end{subfigure}
\hfill
\begin{subfigure}[t]{0.48\textwidth}
    \centering
    \includegraphics[width=\linewidth]{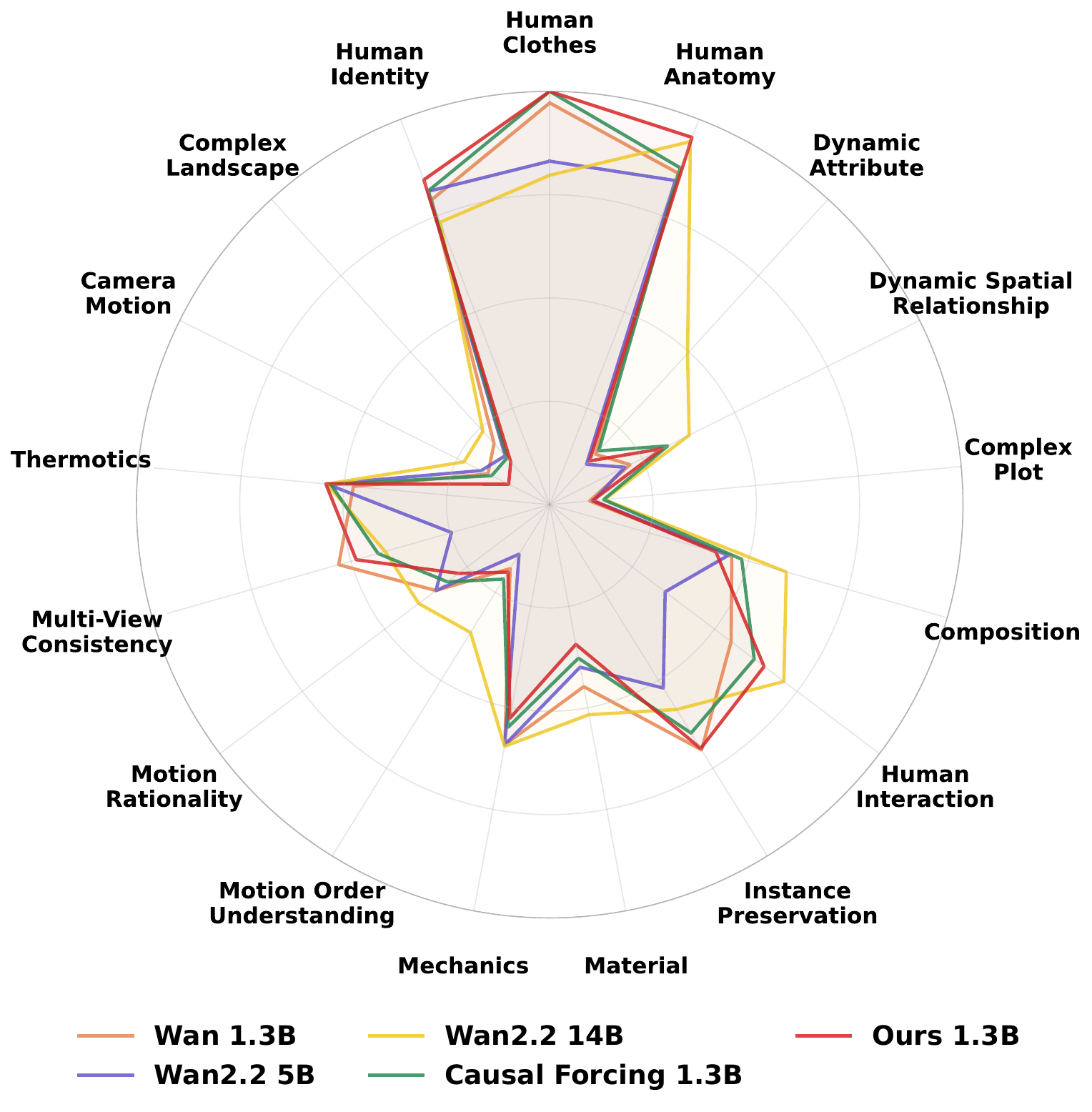}
    \caption{VBench-2.0 per-dimension comparison.}
    \label{fig:vbench2_radar}
\end{subfigure}
\caption{
\textbf{Comparison on general video benchmarks.}
We show normalized radar charts for VBench and VBench-2.0 dimensions. Our model
remains competitive across general perceptual, temporal, and compositional
categories, while showing strong performance on human- and motion-related
dimensions. These results support that \methodName{} improves human-centered
physical plausibility without substantially sacrificing general video generation
quality.
}
\label{fig:vbench_radar_appendix}
\end{figure*}
To assess whether our human-centric reward over-specializes the
generator, we evaluate on the standard VBench and VBench-2.0
prompt sets. We use Causal Forcing 1.3B as the primary comparison,
since it is the backbone of our post-trained model, and include
larger Wan2.2 models as additional references.

As shown in \Cref{tab:general_vbench}, our model maintains
comparable average VBench performance to the base model
($0.654$ vs.\ $0.657$) and slightly improves VBench-2.0
($0.462$ vs.\ $0.458$). These results suggest that optimizing
human-motion feasibility does not substantially compromise
general video generation quality.

In \Cref{fig:vbench_radar_appendix}, the category-level results show that our
model remains competitive across broad perceptual, temporal, and compositional
dimensions, while retaining strong performance on human- and motion-related
categories such as subject consistency, motion smoothness, temporal flickering,
human action, and human anatomy. 

\change{
\section{Reward Computation Overhead}
\label{app:reward_overhead_details}

We compare the standalone reward-evaluation latency and effective
training overhead of \methodName{} against HPSv3~\citep{ma2025hpsv3},
VideoAlign-MQ~\citep{liu2025videoalign}, and
VideoPhy-PC~\citep{bansal2024videophy}.
These baselines evaluate human preference, motion quality,
and physical commonsense, respectively.

\noindent\textbf{Evaluation protocol.}
We generate a fixed batch of $B{=}8$ FastWan videos at
$480{\times}832$ resolution. Each reward is executed once for
warm-up and five times for timing, and we report the steady-state
per-video average.

\noindent\textbf{Effective training overhead.}
Our training pipeline overlaps video sampling with reward
computation: while the model generates the next batch, the
previous batch is evaluated in parallel. Therefore, only the
non-overlapped reward computation contributes to the effective
training overhead, which we measure relative to a
constant-reward baseline:

\begin{equation}
\mathrm{Overhead} =
\frac{
T_{\mathrm{epoch}}^{\mathrm{reward}}
-
T_{\mathrm{epoch}}^{\mathrm{const}}
}{
T_{\mathrm{epoch}}^{\mathrm{const}}
}.
\end{equation}

As reported in \Cref{tab:reward_overhead}, \methodName{} requires
$2.80$s per video for standalone reward computation but introduces
only $7\%$ effective training overhead due to pipelining.
In contrast, HPSv3 introduces $35\%$ overhead despite its
comparable standalone latency. These results demonstrate that
structured 3D motion evaluation can be efficiently integrated
into RL post-training.
\section{Robustness to Human Motion Reconstruction Errors}
\label{app:reconstruction_robustness}

\subsection{Detection of Severe Anatomical Distortions}
\label{app:severe_distortion}
A possible limitation of a reconstruction-based reward is that the body
reconstructor may map visibly distorted human anatomy to a plausible SMPL-X
pose, concealing the visual artifact. To investigate this issue, we first
analyze 3,000 randomly sampled generated videos using Gemini-2.5-Pro; 2.7\%
are identified as containing severe anatomical distortions. We then evaluate
a balanced set of 100 videos from four generators, including 50 severely
distorted and 50 clean clips, with distortion labels agreed upon by three
human raters without access to automatic metric scores.

We measure how reliably each evaluation signal distinguishes the distorted
from the clean group using area under the receiver operating characteristic
curve (AUROC). As shown in Table~\ref{tab:severe_distortion}, the overall
PhyMotion reward achieves an AUROC of 0.913, exceeding VBench Human Anatomy
(0.875), VideoAlign-MQ (0.732), and VideoAlign-VQ (0.587). Each individual
PhyMotion axis also discriminates between the two groups. These findings
suggest that severe visual artifacts frequently leave detectable evidence in
the recovered trajectories, such as implausible joint motion and physical
violations. Nevertheless, reconstruction can still fail to preserve some
image-space defects; our physics-grounded reward complements, rather than
replaces, visual-quality evaluation.

\subsection{Consistency Under Video-to-3D Reconstruction}
\label{app:reconstruction_consistency}
To isolate the effect of reconstruction errors on the reward, we conduct a
controlled motion--video--motion consistency experiment. We apply temporally
smooth perturbations to ground-truth AMASS~\citep{mahmood2019amass} human motions, producing a range
of nearly valid to moderately distorted trajectories. For each of 240
perturbed motion pairs, we compute PhyMotion through two paths: (i) scoring
the perturbed 3D motion directly, and (ii) rendering the same motion into
video, reconstructing the human trajectory using GVHMR, and then scoring the
reconstruction. The two paths share the same underlying motion, allowing us
to quantify how the intermediate video reconstruction changes reward
rankings.

Table~\ref{tab:reconstruction_consistency} reports the Spearman correlation
between direct and reconstructed scores. Correlations reach 0.854 for
kinematics, 0.790 for contact, 0.899 for dynamics, and 0.874 for the overall
reward. This agreement indicates that the relative physical assessments are
largely preserved under the tested reconstruction perturbations. It does not
imply perfect 3D reconstruction or guarantee robustness to severe tracking
failures, unobserved anatomy, or arbitrary scene conditions.

\begin{table}[t]
  \centering
  \caption{Detection of severe anatomical distortion on a balanced set of
  50 distorted and 50 clean generated videos. Higher AUROC is better.}
  \label{tab:severe_distortion}
  \begin{tabular}{lc}
    \toprule
    Metric & AUROC $\uparrow$ \\
    \midrule
    VBench Human Anatomy & 0.875 \\
    VideoAlign-MQ & 0.732 \\
    VideoAlign-VQ & 0.587 \\
    PhyMotion Kinematic & 0.848 \\
    PhyMotion Contact & 0.869 \\
    PhyMotion Dynamic & 0.880 \\
    \textbf{PhyMotion Overall} & \textbf{0.913} \\
    \bottomrule
  \end{tabular}
\end{table}
\begin{table}[t]
  \centering
  \caption{Spearman rank correlation between PhyMotion scores from direct
  3D motions and scores after video rendering and GVHMR reconstruction.}
  \label{tab:reconstruction_consistency}
  \begin{tabular}{lc}
    \toprule
    PhyMotion component & Spearman $\rho\uparrow$ \\
    \midrule
    Kinematic & 0.854 \\
    Contact & 0.790 \\
    Dynamic & 0.899 \\
    \textbf{Overall} & \textbf{0.874} \\
    \bottomrule
  \end{tabular}
\end{table}

\subsection{Challenging Reconstruction Conditions}
\label{app:challenging_cases}
We also inspect representative generated videos involving multiple people,
substantial articulation and self-occlusion, partial-body visibility, and
moving cameras. For multi-person videos, we track and reconstruct each
person separately and aggregate the resulting per-person rewards using a
weighted average. Occluded joints are estimated using temporal evidence and
the SMPL-X body prior; this can provide a coherent trajectory under moderate
occlusion, but also creates a risk of concealing visual defects. Changes in
camera pose further require a consistent reference frame to distinguish
body motion from viewpoint changes. Full temporal examples include group
dancing, martial-arts actions, skating, inversion, and floor exercises.
These qualitative demonstrations illustrate the scope of the pipeline but
do not establish reliability under complete occlusions or lost tracks.

\section{Extending PhyMotion to Non-Flat Terrain}
\label{app:terrain_robustness}

Our default contact and balance evaluation assumes a flat ground plane, which may incorrectly penalize valid contacts on non-flat terrain. To extend PhyMotion, we incorporate terrain geometry into the MuJoCo simulator and evaluate foot contact, penetration, and balance relative to local support surfaces rather than a single global plane. This requires only modifying the support geometry and contact evaluation, while preserving the overall reward formulation.

We validate this extension on six ground-truth stair-climbing motions from AMASS~\citep{mahmood2019amass}. Incorporating local staircase geometry improves the average contact score from 0.75 to 0.90. In a controlled synthetic ascent, valid-contact recognition increases from 0.25 to 0.94. Furthermore, a sim-to-real-to-sim experiment shows that GVHMR recovers all five staircase levels and 0.73\,m of vertical displacement from a photorealistic video, compared with 0.81\,m in the original motion. These results demonstrate that PhyMotion can accommodate non-flat terrain with minimal modifications.

\section{Analysis of Action Execution and Motion Pacing}
\label{app:motion_suppression}

\subsection{Instruction-Conditioned Action Execution}
\label{app:instruction_execution}
The decrease in VBench Dynamic Degree following RL post-training raises
the question of whether the generator avoids physically difficult actions
in order to earn higher feasibility rewards. To test this, we use Gemini
to evaluate anonymized generated videos conditioned on their text prompts,
classifying the requested action as fully executed, partially executed,
or not executed. Table~\ref{tab:instruction_execution} reports the
results, with larger Wan models included as references rather than
controlled before/after comparisons.

Relative to the same Causal Forcing 1.3B backbone before RL, PhyMotion
post-training increases the full-execution rate from 13.9\% to 21.5\%
(+7.6 percentage points). Full-or-partial execution increases from 54.3\%
to 57.8\%, while non-execution decreases from 45.7\% to 42.3\%. This
provides evidence against broad action avoidance as the explanation for
improved rewards, while leaving open the possibility of more conservative
motion timing.

\begin{table}[t]
  \centering
  \caption{Prompt-conditioned action execution judged by Gemini (\%).}
  \label{tab:instruction_execution}
  \begin{tabular}{lrrrr}
    \toprule
    Generator & Full $\uparrow$ & Partial & None $\downarrow$ & Full+Partial $\uparrow$ \\
    \midrule
    Causal Forcing 1.3B & 13.9 & 40.4 & 45.7 & 54.3 \\
    PhyMotion RL (1.3B) & \textbf{21.5} & 36.3 & \textbf{42.3} & \textbf{57.8} \\
    \midrule
    Wan2.2 5B & 11.4 & 43.0 & 45.6 & 54.4 \\
    Wan2.2 14B & 34.4 & 36.3 & 29.3 & 70.7 \\
    \bottomrule
  \end{tabular}
\end{table}

\subsection{Human Evaluation of Motion Pacing}
\label{app:human_pacing}
We separately evaluate perceived pacing on 500 generated videos, equally
split between the Causal Forcing baseline and the PhyMotion-finetuned
model (250 per model). Three human annotators judge whether each video's
motion is overly fast or jittery, appropriately paced for the requested
action, or overly slow or insufficiently dynamic. As reported in
Table~\ref{tab:human_pacing}, the share of normally paced clips is nearly
unchanged after RL (88\% to 87\%), and overly fast or jittery motion
falls from 10\% to 4\%. However, the overly slow category rises from
3\% to 8\%. Thus, the results do not indicate widespread collapse to
static videos, but they do reveal a modest slowdown trade-off. Combining
physical feasibility rewards with complementary action-completion or
motion-amplitude objectives is a possible direction for addressing this
limitation.

\begin{table}[t]
  \centering
  \caption{Human judgments of motion pacing on 500 videos, with three
  raters.}
  \label{tab:human_pacing}
  \begin{tabular}{lccc}
    \toprule
    Generator & Overly fast (\%) & Normal (\%) & Overly slow (\%) \\
    \midrule
    Causal Forcing 1.3B & 10 & 88 & 3 \\
    PhyMotion RL (1.3B) & 4 & 87 & 8 \\
    \bottomrule
  \end{tabular}
\end{table}

}

\section{Human Study Detail}
\label{app:human_detail}

This appendix describes the human preference studies used to evaluate whether
\methodName{} aligns with human judgments of generated human-motion quality. We
conduct two pairwise-comparison studies: one for validating our automatic
feasibility metrics against human preferences, and one for ranking post-trained
and baseline video generation models. Both studies were conducted with non-author 
participants, including students and researchers in related areas. Participants compared short generated videos under the same text
prompt and judged them along three axes: body structure, balance, and motion
naturalness.

\begin{figure}[h]
    \centering
    \includegraphics[width=\linewidth]{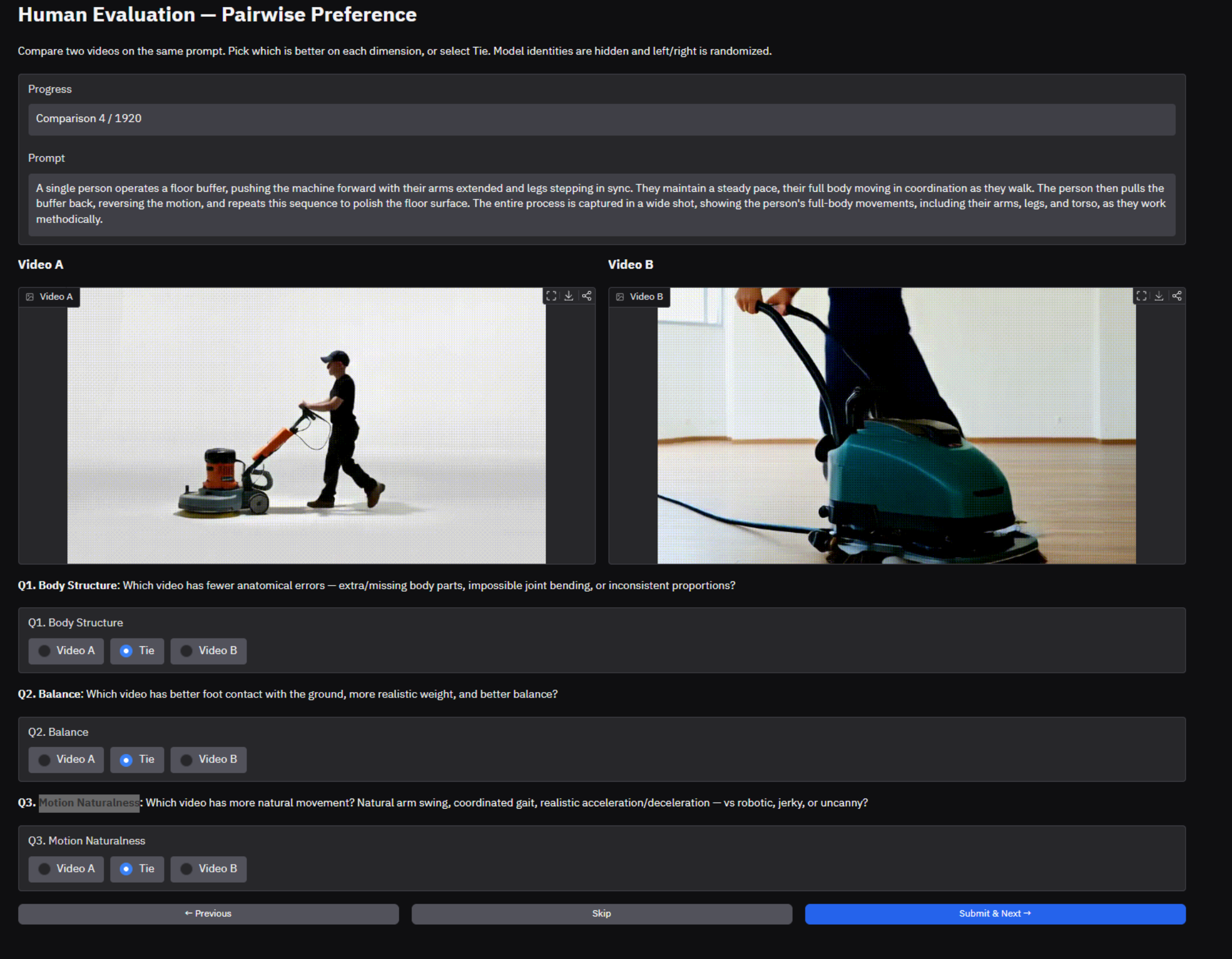}
    \caption{
    \textbf{Human preference annotation interface.}
  Annotators compare two anonymized videos generated from the same prompt and
  answer three pairwise preference questions on body structure, balance, and
  motion naturalness.
    }
    \label{fig:study-ui}
\end{figure}

\subsection{Task Design}
\label{app:study-task}

Each annotation example consists of two short videos generated from the same
text prompt by two different models. The videos are displayed side by side and
played in sync. Model identities are hidden from annotators, and the left/right
order is randomized independently for each pair.

For each video pair, annotators answer three preference questions, corresponding
to the main perceptual dimensions studied in this paper:

\begin{itemize}
    \item \textbf{Body structure:} Which video shows a more anatomically correct
    human body? Annotators are asked to consider limb consistency, body-part
    distortions, missing or extra limbs, and self-intersections.

    \item \textbf{Balance:} Which video shows more physically plausible body
    support? Annotators are asked to consider whether the person appears stably
    supported by the contacting feet, hands, or other body parts.

    \item \textbf{Motion naturalness:} Which video shows motion that looks more
    like a real human action? Annotators are asked to consider temporal
    smoothness, realistic timing, and the absence of teleporting, flickering, or
    floating artifacts.
\end{itemize}

For each question, annotators choose one of three options:
\textsf{Video A}, \textsf{Video B}, or \textsf{Tie}. The default option is
\textsf{Tie}, so unanswered questions do not introduce a preference toward
either side. The same instruction text is shown to every annotator, and a
screenshot of the annotation interface is shown in~\Cref{fig:study-ui}.

\subsection{Study Composition}
\label{app:study-rounds}

We form the dataset using a wide variety of video models
including Causal Forcing 1.3B~\citep{zhu2026causal}, FastWan 1.3B~\citep{zhang2025fast}, Wan 2.1 1.3B, Wan2.2 5B, Wan2.2
14B~\citep{wan2025wan}, and EchoMotion 5B~\citep{yang2025echomotion}. Each pair contains two videos generated from the same
prompt by different models. We filter out unsafe videos, nearly identical pairs,
and pairs where both videos already have very high feasibility scores, since
these provide little discriminative signal. After filtering incomplete sessions
and low-coverage variants, the final Elo analysis uses $1{,}487$ vote rows per
question, or $4{,}461$ axis-level annotations in total. Ties are counted as
$0.5$, and Elo ratings are computed with base rating $1500$ and $K=32$.

\subsection{Inter-Annotator Agreement}
We also compute inter-annotator agreement on overlapping video pairs in the
pilot study. Since some comparisons are ambiguous, we treat \textsf{Tie} as
compatible with either \textsf{Video A} or \textsf{Video B}; a disagreement is
counted only when two annotators make hard opposite choices, i.e., one selects
\textsf{Video A} and the other selects \textsf{Video B}. As shown in
\Cref{tab:pilot_agreement}, annotators achieve high agreement across all three
questions, suggesting that the criteria are consistently understood.
\begin{table}[t]
\centering
\caption{
\textbf{Inter-annotator agreement on the human pilot study.}
For each question, we report the mean pairwise agreement between human
annotators on overlapping video pairs.
}
\label{tab:pilot_agreement}
\begin{tabular}{lccc}
\toprule
Question & Body Structure & Balance & Motion Naturalness \\
\midrule
Mean Pairwise Agreement & 95.9\% & 91.4\% & 96.9\% \\
\bottomrule
\end{tabular}
\end{table}

\subsection{Human Study Significance}
\label{app:human_sig}
\begin{table*}[t]
\centering
\scriptsize
\setlength{\tabcolsep}{4.0pt}
\renewcommand{\arraystretch}{1.08}
\caption{
\textbf{Correlation with human judgments and statistical significance.}
We report Spearman's rank correlation ($\rho$) between automatic metric scores
and human judgments across three questions, together with significance levels.
Best results are in \textbf{bold}, and second-best results are
\underline{underlined}.
}
\label{tab:human_alignment_sigma}
\resizebox{\textwidth}{!}{
\begin{tabular}{llcccc}
\toprule
\textbf{Metric Group} & \textbf{Metric}
& \textbf{Body Structure}
& \textbf{Balance}
& \textbf{Motion Naturalness}
& \textbf{All Questions} \\
\midrule

\multirow{5}{*}{VBench / VBench2}
& Aesthetic
& $+0.248 \pm 0.032$ & $+0.238 \pm 0.033$ & $+0.260 \pm 0.032$ & $+0.248 \pm 0.018$ \\
& Motion
& $+0.236 \pm 0.035$ & $+0.246 \pm 0.034$ & $+0.274 \pm 0.034$ & $+0.252 \pm 0.019$ \\
& Temporal
& $+0.175 \pm 0.032$ & $+0.171 \pm 0.035$ & $+0.208 \pm 0.034$ & $+0.185 \pm 0.020$ \\
& Dynamic
& $-0.135 \pm 0.033$ & $-0.118 \pm 0.035$ & $-0.149 \pm 0.034$ & $-0.135 \pm 0.019$ \\
& Anatomy
& $+0.273 \pm 0.032$ & $+0.235 \pm 0.033$ & $+0.278 \pm 0.033$ & $+0.262 \pm 0.018$ \\
\midrule

\multirow{2}{*}{VideoAlign}
& Motion
& $+0.155 \pm 0.035$ & $+0.175 \pm 0.036$ & $+0.156 \pm 0.034$ & $+0.161 \pm 0.020$ \\
& Visual
& $+0.045 \pm 0.035$ & $+0.053 \pm 0.035$ & $+0.067 \pm 0.034$ & $+0.055 \pm 0.020$ \\
\midrule

VideoPhy
& Physics
& $+0.168 \pm 0.033$ & $+0.142 \pm 0.035$ & $+0.108 \pm 0.035$ & $+0.138 \pm 0.020$ \\
\midrule

\multirow{4}{*}{\methodName (Ours)}
& Kinematic
& $\underline{+0.369 \pm 0.031}$ & $+0.314 \pm 0.033$ & $+0.375 \pm 0.031$ & $+0.353 \pm 0.018$ \\
& Contact
& $+0.290 \pm 0.032$ & $\mathbf{+0.337 \pm 0.033}$ & $+0.281 \pm 0.032$ & $+0.290 \pm 0.018$ \\
& Dynamic
& $+0.367 \pm 0.030$ & $+0.316 \pm 0.032$ & $\underline{+0.389 \pm 0.030}$ & $\underline{+0.358 \pm 0.017}$ \\
& Overall
& $\mathbf{+0.391 \pm 0.030}$ & $\underline{+0.333 \pm 0.032}$ & $\mathbf{+0.402 \pm 0.030}$ & $\mathbf{+0.376 \pm 0.017}$ \\

\bottomrule
\end{tabular}
}
\end{table*}
\paragraph{Correlation with human judgments.} In addition to the correlation values reported in the main paper, we report the
statistical significance of the metric--human alignment results in ~\Cref{tab:human_alignment_sigma}.
\begin{table}[t]
\centering
\caption{
\textbf{Detailed human preference Elo ratings.}
We provide the full Elo ratings corresponding to the human preference evaluation
summarized in the main paper. Ratings are computed from pairwise human
preferences on body structure, balance, motion naturalness, and all questions.
Higher is better, and uncertainty denotes bootstrap standard deviation.
}
\label{tab:human_preference_appendix}
\footnotesize
\setlength{\tabcolsep}{4.5pt}
\renewcommand{\arraystretch}{1.05}
\begin{tabular}{lcccc}
\toprule
Model & Body Struct. & Balance & Motion & Overall \\
\midrule
\textbf{Ours}
& $\mathbf{1620 \pm 34}$
& $\mathbf{1610 \pm 34}$
& $\mathbf{1632 \pm 39}$
& $\mathbf{1621 \pm 11}$ \\
Wan2.2 14B
& $\underline{1600 \pm 34}$
& $\underline{1593 \pm 33}$
& $\underline{1618 \pm 40}$
& $\underline{1604 \pm 13}$ \\
Causal-Forcing 1.3B
& $1562 \pm 35$
& $1546 \pm 33$
& $1553 \pm 39$
& $1553 \pm 8$ \\
FastWan 1.3B
& $1526 \pm 33$
& $1521 \pm 32$
& $1528 \pm 35$
& $1525 \pm 4$ \\
Wan 1.3B
& $1429 \pm 37$
& $1440 \pm 33$
& $1411 \pm 39$
& $1427 \pm 15$ \\
EchoMotion 5B
& $1386 \pm 36$
& $1403 \pm 35$
& $1374 \pm 38$
& $1387 \pm 14$ \\
Wan2.2 5B
& $1376 \pm 35$
& $1388 \pm 34$
& $1384 \pm 39$
& $1383 \pm 6$ \\
\bottomrule
\end{tabular}
\end{table}
\paragraph{Human preference evaluation.}
To complement the compact human preference results in the main paper, we report
the full per-question Elo ratings with bootstrap uncertainty in
\Cref{tab:human_preference_appendix}.

\subsection{Annotator Recruitment, Compensation, and Data Handling.}
Annotators were recruited from paid research assistants
within our laboratory, and annotation was conducted as
part of their regular research responsibilities.
After data collection, personal identifying information
was removed from the annotation records, and only
de-identified annotator IDs (e.g., Annotator 1, Annotator 2)
were retained for subsequent analysis.

\section{Detailed Definition of \methodName{} Metrics}
\label{app:metric_details}

This appendix provides the detailed computation of the three \methodName{}
feasibility scores used in the main paper: kinematic feasibility, contact
feasibility, and dynamic feasibility. The goal of these metrics is to convert a
generated human video into interpretable physical signals that can be used both
for evaluation and as rewards for RL post-training.

Given a generated video $v=\{I_t\}_{t=1}^T$ with frame rate $f$, we first recover
an SMPL-X body trajectory using GVHMR. Let $q_t$ denote the recovered body pose
at frame $t$, $X_t=\{x_{t,j}\}_{j=1}^{J}$ denote the 3D body joints, and $M_t$
denote the body mesh. For SMPL-X, we use $J=55$ joints and a mesh with
$V=10{,}475$ vertices and $F=20{,}908$ triangular faces. We then retarget the
recovered motion to a MuJoCo human model and compute physics-related quantities
such as joint torques and ground reaction forces. We write $\Delta t=1/f$.

For foot-related metrics, we define two foot-sole vertex sets,
$\mathcal{V}_L$ and $\mathcal{V}_R$, corresponding to the left and right feet.
These sets are precomputed from the SMPL-X template by selecting the lowest
vertices on each foot. In our experiments, all metrics are computed from the
GVHMR-recovered camera-frame trajectory, with $f=16$.

\subsection{Kinematic Feasibility}

Kinematic feasibility measures whether the reconstructed human body is
anatomically valid and temporally smooth. It penalizes three common artifacts:
joints moving too fast, body parts intersecting each other, and unstable limb
geometry. The final kinematic score is high when the motion has smooth joints,
little self-penetration, and a stable body structure.

\paragraph{Joint angular velocity.}
Generated videos sometimes contain abrupt limb snapping, where a joint rotates
too quickly between adjacent frames. We approximate the angular velocity of
joint $j$ at frame $t$ by finite differences:
$$
\omega_{t,j} = f \, d_{\mathcal{Q}}(q_{t+1,j}, q_{t,j}),
$$
where $d_{\mathcal{Q}}(\cdot,\cdot)$ measures the distance between two joint
rotations, such as the geodesic distance on $SO(3)$. We then compute the
fraction of joint-frame pairs whose angular velocity exceeds a plausible
joint-specific threshold:
$$
v_{\mathrm{vel}}
=
\frac{1}{(T-1)J}
\sum_{t=1}^{T-1}
\sum_{j=1}^{J}
\mathbf{1}
\left[
\omega_{t,j} > \omega^{\max}_{j}
\right].
$$

\paragraph{Self-penetration.}
Another common failure is body-part penetration, such as a hand passing through
the torso or a leg intersecting the body. For each frame, we detect intersections
between non-identical mesh triangles using a BVH-based triangle-intersection
test. Let $\mathcal{F}$ be the set of mesh triangles. We compute the percentage
of intersecting triangle pairs:
$$
\mathrm{spen}_t
=
100 \cdot
\frac{
\left|
\left\{
(f_1,f_2)\in\mathcal{F}^2 :
f_1 \cap f_2 \neq \emptyset,\;
f_1 \neq f_2
\right\}
\right|
}{F}.
$$

We then average this quantity over time:
$$
\mathrm{spen}
=
\frac{1}{T}
\sum_{t=1}^{T}
\mathrm{spen}_t.
$$

Because clean SMPL meshes can have small soft-body overlaps, we treat roughly
$2\%$ self-penetration as a normal baseline and regard values above $20\%$ as severe.
The normalized self-penetration violation is
$$
v_{\mathrm{spen}}
=
\mathrm{clip}
\left(
\frac{\mathrm{spen}-2}{18},
0,
1
\right).
$$

\paragraph{Joint-limit violation.}
Generated videos may contain anatomically impossible poses, such as a knee
bending backward, a hip twisting too far, or an arm rotating beyond a plausible
range. To detect these failures, we compare each reconstructed joint angle with
the valid joint range defined by the MuJoCo human model. Let
$q_{t,j}$ denote the pose parameter of joint $j$ at frame $t$, and let
$[q_j^{\min}, q_j^{\max}]$ denote its valid range. We define the joint-limit
violation as the fraction of joint-frame pairs that fall outside this range:
$$
v_{\mathrm{lim}}
=
\frac{1}{TJ}
\sum_{t=1}^{T}
\sum_{j=1}^{J}
\mathbf{1}
\left[
q_{t,j} < q_j^{\min}
\;\vee\;
q_{t,j} > q_j^{\max}
\right].
$$

For multi-dimensional joints, the violation is computed per degree of freedom
and then averaged across all joint dimensions. This term penalizes poses that
may look locally plausible in RGB frames but require anatomically invalid joint
configurations in 3D.

Finally, the kinematic feasibility score is
$$
F_{\mathrm{kin}}(v)
=
1 -
\frac{1}{3}
\left(
v_{\mathrm{vel}}
+
v_{\mathrm{spen}}
+
v_{\mathrm{lim}}
\right).
$$

Higher $F_{\mathrm{kin}}$ indicates smoother and more anatomically plausible
body motion.

\subsection{Contact Feasibility}

Contact feasibility measures whether the body interacts with the ground in a
physically plausible way. It penalizes foot sliding, ground penetration, floating
feet, and poor balance.

For each foot $k\in\{L,R\}$, let $u_{t,v}$ denote the 3D position of mesh vertex
$v$ at frame $t$. We compute the foot height using the lowest sole vertex:
$$
h_{t,f}
=
\min_{v\in\mathcal{V}_f}
u_{t,v}^{(z)}.
$$

We also compute the average foot-sole position:
$$
p_{t,f}
=
\frac{1}{|\mathcal{V}_f|}
\sum_{v\in\mathcal{V}_f}
u_{t,v}.
$$

A foot is considered to be in contact with the ground if it is both close to the
ground and nearly stationary:
$$
c_{t,k}
=
\mathbf{1}
\left[
h_{t,f}<0.02
\;\wedge\;
\|\dot{p}_{t,f}\|_2 < 0.05
\right].
$$

\paragraph{Foot sliding.}
If a foot is in contact with the ground, it should not slide significantly. We
therefore measure the amount of foot displacement while contact is active:
$$
\mathrm{slip}_{t,f}
=
c_{t,k}
\cdot
\|\dot{p}_{t,f}\|_2
\cdot
\Delta t.
$$

The average foot-sliding violation is
$$
v_{\mathrm{slip}}
=
\frac{1}{2T}
\sum_{t=1}^{T}
\sum_{k\in\{L,R\}}
\mathrm{slip}_{t,f}.
$$

\paragraph{Ground penetration.}
Generated or reconstructed feet can sometimes go below the ground plane. We
measure penetration depth as
$$
\mathrm{gpen}_{t,f}
=
\max(0,-h_{t,f}),
$$
and average it over both feet and all frames:
$$
v_{\mathrm{gpen}}
=
\frac{1}{2T}
\sum_{t=1}^{T}
\sum_{k\in\{L,R\}}
\mathrm{gpen}_{t,f}.
$$

\paragraph{Foot floating.}
A common artifact is that the root body moves while the feet do not make
plausible contact with the ground, producing floating or whipping feet. To
capture this, we compare the foot motion relative to the root with the root
motion itself:
$$
\rho_{t,f}
=
\frac{
\|\dot{(p_{t,f}-x_{t,0})}\|_2
}{
\|\dot{x}_{t,0}\|_2+\varepsilon
},
$$
where $x_{t,0}$ denotes the pelvis/root joint. We flag a frame if the foot
moves implausibly slowly relative to the root, or implausibly fast:
$$
\rho_{t,f}<0.6
\quad\text{or}\quad
\rho_{t,f}>1.75.
$$

The floating violation is the fraction of frames where such implausible
foot-root motion is detected:
$$
v_{\mathrm{float}}
=
\frac{1}{2T}
\sum_{t=1}^{T}
\sum_{k\in\{L,R\}}
\mathbf{1}
\left[
\rho_{t,f}<0.6
\;\vee\;
\rho_{t,f}>1.75
\right].
$$

In implementation, we additionally include sequence-level checks for sustained
non-contact and whole-body floating, which help detect cases where both feet
remain airborne with an irregular trajectory.

\paragraph{Balance.}
A physically plausible standing or walking person should keep the projected
center of mass close to the support region formed by the contacting feet. We
approximate the center of mass as a weighted average of body joints, giving the
pelvis three times the weight of other joints:
$$
C_t
=
\frac{1}{\sum_j w_j}
\sum_{j=0}^{J-1}
w_j x_{t,j},
\qquad
w_0=3,\quad
w_{j>0}=1.
$$

Let $\mathcal{P}_t$ denote the support polygon, defined as the convex hull of
the contacting ankle positions in the ground plane. We compute the distance from
the projected center of mass $C_t^{(xy)}$ to this support polygon:
$$
d_t
=
\begin{cases}
1.0, & \text{if there is no support contact},\\
\min_{e\in\partial\mathcal{P}_t}
\mathrm{dist}_e(C_t^{(xy)}), & \text{otherwise}.
\end{cases}
$$

The balance violation is the normalized average distance:
$$
v_{\mathrm{bal}}
=
\frac{1}{T}
\sum_{t=1}^{T}
\frac{
\mathrm{clip}(d_t,0,0.5)
}{0.5}.
$$

We use this continuous balance measure instead of a binary stable/unstable
indicator because binary thresholds often saturate and provide less useful
training signal.

The final contact feasibility score is
$$
F_{\mathrm{con}}(v)
=
1 -
\frac{1}{4}
\left(
v_{\mathrm{slip}}
+
v_{\mathrm{gpen}}
+
v_{\mathrm{float}}
+
v_{\mathrm{bal}}
\right),
$$
where $v_{\mathrm{slip}}$ and $v_{\mathrm{gpen}}$ are normalized
to $[0,1]$ using fixed thresholds. Higher $F_{\mathrm{con}}$ indicates more
plausible body-ground interaction.

\subsection{Dynamic Feasibility}

Dynamic feasibility measures whether the recovered motion could be produced by a
physically plausible human body. It penalizes motions that require excessive
ground reaction forces, excessive joint torques, or unusually large mechanical
effort.

\paragraph{Ground reaction force.}
We estimate the ground reaction force needed to produce the observed center-of-mass acceleration. Let $m=70\,\mathrm{kg}$ be the body mass and
$g=9.81\,\mathrm{m/s^2}$ be gravity. Using Newton's second law, we estimate
$$
F^{\mathrm{GRF}}_t
=
\begin{bmatrix}
m\ddot{C}_t^{(x)}\\
m\ddot{C}_t^{(y)}\\
m(g+\ddot{C}_t^{(z)})
\end{bmatrix}.
$$

We penalize vertical forces larger than $3$ times body weight:
$$
v_{\mathrm{grf}}^{\mathrm{vert}}
=
\frac{1}{T}
\sum_{t=1}^{T}
\mathbf{1}
\left[
(F^{\mathrm{GRF}}_t)^{(z)} > 3mg
\right].
$$

We also penalize horizontal forces larger than $0.5$ times body weight:
$$
v_{\mathrm{grf}}^{\mathrm{horiz}}
=
\frac{1}{T}
\sum_{t=1}^{T}
\mathbf{1}
\left[
\sqrt{
(F^{\mathrm{GRF}}_t)^{(x)2}
+
(F^{\mathrm{GRF}}_t)^{(y)2}
}
>
0.5mg
\right].
$$

The ground-reaction-force score is
$$
s_{\mathrm{GRF}}
=
1 -
\frac{1}{2}
\left(
v_{\mathrm{grf}}^{\mathrm{vert}}
+
v_{\mathrm{grf}}^{\mathrm{horiz}}
\right).
$$

\paragraph{Joint torque.}
We estimate the torque required to drive each joint using a simple segment-level
inertia model:
$$
\tau_{t,j}
=
I_j \|\ddot{x}_{t,j}\|_2,
$$
where $I_j$ is the approximate inertia for joint $j$. In our implementation, we
use $I_j=1.0\,\mathrm{kg\,m^2}$ for all joints.

For each joint type, we compute the fraction of frames where the estimated
torque exceeds a plausible limit:
$$
v_{\mathrm{torque}}^{(j)}
=
\frac{1}{T}
\sum_{t=1}^{T}
\mathbf{1}
\left[
\tau_{t,j} > \tau_{\max}^{(j)}
\right].
$$

We use torque limits of $200$, $300$, $400$, and $200\,\mathrm{N\,m}$ for the
ankle, knee, hip, and spine, respectively. The torque score is
$$
s_{\tau}
=
1 -
\frac{1}{J}
\sum_{j=1}^{J}
v_{\mathrm{torque}}^{(j)}.
$$

\paragraph{Mechanical effort.}
Even if a motion does not exceed a single force or torque threshold, it may
still require unrealistically large total effort. We therefore compute a
mechanical-work proxy by multiplying joint torque with joint velocity:
$$
\mathrm{MET}
=
\sum_{t=1}^{T}
\sum_{j=0}^{J-1}
\tau_{t,j}
\|\dot{x}_{t,j}\|_2
\Delta t.
$$

We convert this value into a normalized score:
$$
s_{\mathrm{met}}
=
\max
\left(
0,
1-\frac{\mathrm{MET}}{10000}
\right).
$$

The final dynamic feasibility score is
$$
F_{\mathrm{dyn}}(v)
=
\frac{1}{3}
\left(
s_{\tau}
+
s_{\mathrm{GRF}}
+
s_{\mathrm{met}}
\right).
$$

Higher $F_{\mathrm{dyn}}$ indicates that the motion requires more plausible
forces, torques, and total effort.

\subsection{Overall Reward}

The final \methodName{} reward is the average of the three feasibility axes:
$$
R_{\mathrm{motion}}(v)
=
\frac{1}{3}
\left(
F_{\mathrm{kin}}(v)
+
F_{\mathrm{con}}(v)
+
F_{\mathrm{dyn}}(v)
\right).
$$

All violation terms are clipped or normalized to $[0,1]$, so higher scores
consistently indicate more physically plausible human motion. This decomposition
also makes the reward interpretable: kinematic feasibility captures body-shape
and joint-motion artifacts, contact feasibility captures body-ground interaction
failures, and dynamic feasibility captures physically infeasible forces and
motions.
\begin{table}[ht]
\centering
\small
\caption{Training hyperparameters for \methodName{} training on Causal Forcing-1.3B  and FastWan-1.3B. For the training time of FastWan, part of the GPU was running another job, so the reported time is longer than expected.}
\setlength{\tabcolsep}{6pt}
\begin{tabular}{lll}
\toprule
\textbf{Setting} & \textbf{Causal Forcing-1.3B } & \textbf{FastWan-1.3B} \\
\midrule
\multicolumn{3}{l}{\emph{Backbone \& reward}} \\
Base model              & Causal Forcing-1.3B (chunkwise) & FastWan-1.3B (bidirectional) \\
Dataset                 & MotionX ($21{,}348$ training prompts) & MotionX ($21{,}348$ training prompts) \\
Resolution / frames     & $480 \times 832$, $45$ frames & $480 \times 832$, $45$ frames \\
\midrule
\multicolumn{3}{l}{\emph{LoRA}} \\
Rank $r$                & $256$ & $256$ \\
Scale $\alpha$          & $256$ & $256$ \\
Dropout                 & $0$ & $0$ \\
\midrule
\multicolumn{3}{l}{\emph{Optimization}} \\
Optimizer               & AdamW & AdamW \\
Learning rate           & $1\times10^{-5}$ & $1\times10^{-5}$ \\
$(\beta_1, \beta_2)$    & $(0.9,\ 0.999)$ & $(0.9,\ 0.999)$ \\
Adam $\epsilon$         & $1\times10^{-8}$ & $1\times10^{-8}$ \\
Weight decay            & $1\times10^{-4}$ & $1\times10^{-4}$ \\
Gradient clipping       & $\|g\|_2 \le 1.0$ & $\|g\|_2 \le 1.0$ \\
Mixed precision         & bf16 & bf16 \\
\midrule
\multicolumn{3}{l}{\emph{NFT objective}} \\
KL interpolation $\beta$        & $0.1$ & $0.1$ \\
KL coefficient $\lambda$        & $1\times10^{-4}$ & $1\times10^{-4}$ \\
Advantage normalization         & per-prompt, $|A|\le 5$ & per-prompt, $|A|\le 5$ \\
\midrule
\multicolumn{3}{l}{\emph{Rollout / sampling}} \\
Denoising step list             & $\{1000, 750, 500, 250\}$ & $\{999, 856, 599\}$ \\
\texttt{num\_frame\_per\_block} & $3$ & N/A (bidirectional, full-sequence) \\
KV-cache                        & enabled & n/a (bidirectional) \\
Guidance scale                  & $1.0$ (post-distillation) & $1.0$ (post-distillation) \\
Samples per prompt              & $24$ & $24$ \\
Rollout batches per epoch       & $16$ & $16$ \\
\midrule
\multicolumn{3}{l}{\emph{EMA}} \\
EMA on LoRA params              & yes & yes \\
EMA start step                  & $0$ & $0$ \\
\midrule
\multicolumn{3}{l}{\emph{Schedule \& hardware}} \\
Per-rank micro-batch size       & $3$ & $3$ \\
Gradient accumulation steps     & $16$ & $16$ \\
Reported step                   & $330$ & $330$ \\
GPUs                            & $8\times$ A100 80\,GB & $8\times$ A100 80\,GB \\
Wall-clock                      & $\sim 60.9$ hours & $\sim$ 66 hours (crowded GPU) \\
\bottomrule
\end{tabular}
\label{tab:s210_hparams}
\end{table}

\section{Training Details}
\label{app:training_detail}

We provide the full training and rollout configuration used for our method in
\Cref{tab:s210_hparams}. Unless otherwise specified, all experiments
use the same configuration. We report the checkpoint at training step $330$.

\section{Licenses for Existing Assets}
\label{app:license}
\begin{table}[t]
\centering
\small
\setlength{\tabcolsep}{4pt}
\caption{Licenses of external assets, models, datasets, codebases, and evaluation tools used in this work.}
\begin{tabular}{p{0.32\linewidth}p{0.36\linewidth}p{0.22\linewidth}}
\toprule
Asset / Method & Website / Source & License \\
\midrule
Wan series~\citep{wan2025wan} & \url{https://github.com/Wan-Video/Wan2.2} & Apache-2.0 \\
Causal Forcing~\citep{zhu2026causal} & \url{https://github.com/thu-ml/Causal-Forcing} & Apache-2.0 \\
FastWan~\citep{zhang2025fast} & \url{https://github.com/hao-ai-lab/fastvideo} & Apache-2.0 \\
EchoMotion~\citep{yang2025echomotion} & \url{https://github.com/D2I-ai/EchoMotion} & CC BY-NC 4.0 \\
VideoAlign~\citep{liu2025videoalign} & \url{https://github.com/KlingAIResearch/VideoAlign} & MIT \\
VideoPhy~\citep{bansal2024videophy} & \url{https://github.com/Hritikbansal/videophy} & MIT \\
GVHMR~\citep{shen2024gvhmr} & \url{https://github.com/zju3dv/GVHMR} & Educational, research, and non-profit use only \\
Motion-X~\citep{lin2023motionx} & \url{https://github.com/IDEA-Research/Motion-X} & Non-commercial scientific research only \\
VBench / VBench-2.0~\citep{huang2024vbench,zheng2025vbench2} & \url{https://github.com/Vchitect/VBench} & Apache-2.0 \\
Astrolabe~\citep{zhang2026astrolabe} & \url{https://github.com/franklinz233/Astrolabe} & Apache-2.0 \\
SMPL~\citep{loper2015smpl} & \url{https://smpl.is.tue.mpg.de/} & SMPL-Body Creative Commons License / CC BY 4.0 \\
MuJoCo~\citep{todorov2012mujoco} & \url{https://github.com/google-deepmind/mujoco} & Apache-2.0 \\
\bottomrule
\end{tabular}
\label{tab:asset_licenses}
\end{table}
We show the licenses of all assets we use in \Cref{tab:asset_licenses}

\end{document}